\documentclass[10pt,twocolumn,letterpaper]{article}

\usepackage[pagenumbers]{wacv} 

\usepackage{graphicx}
\usepackage{amsmath}
\usepackage{amssymb}
\usepackage{booktabs}
\usepackage{svg}
\usepackage{booktabs}
\usepackage{multirow}
\usepackage{siunitx}
\usepackage{float} 

\usepackage{xr}
\makeatletter

\newcommand*{\addFileDependency}[1]{
\typeout{(#1)}
\@addtofilelist{#1}
\IfFileExists{#1}{}{\typeout{No file #1.}}
}\makeatother

\newcommand*{\myexternaldocument}[1]{%
\externaldocument{#1}%
\addFileDependency{#1.tex}%
\addFileDependency{#1.aux}%
}

\myexternaldocument{Suppl}

\usepackage[pagebackref,breaklinks,colorlinks]{hyperref}

\usepackage[%
  xindy,
  acronym,
]{glossaries}
\glsdisablehyper

\usepackage[capitalize]{cleveref}
\crefname{section}{Sec.}{Secs.}
\Crefname{section}{Section}{Sections}
\Crefname{table}{Table}{Tables}
\crefname{table}{Tab.}{Tabs.}

\def\wacvPaperID{2354} 
\def\confName{WACV}
\def\confYear{2025}

\newacronym{gl:GAN}{GAN}{Generative Adversarial Network}
\newacronym{gl:CNN}{CNN}{Convolutional Neural Network}
\newacronym{gl:SR}{SR}{Super-resolution}
\newacronym{gl:MLP}{MLP}{Multi-Layer Perceptron}
\newacronym{gl:RRDB}{RRDB}{Residual-in-Residual Dense Block}
\newacronym{gl:OSM}{OSM}{OpenStreetMap}
\newacronym{gl:TCI}{TCI}{True Color Images} 
\newacronym{gl:UTM}{UTM}{Universal Transverse Mercator}

\usepackage{booktabs}
\usepackage{array}
\newcolumntype{x}{l}
\newcolumntype{X}{>{\scriptsize}l}
\newcolumntype{v}[1]{>{\raggedright\hspace{0pt}}p{#1}}
\newcolumntype{V}[1]{>{\scriptsize\raggedright\hspace{0pt}}p{#1}}

\begin{document}

\title{Can Location Embeddings Enhance Super-Resolution of Satellite Imagery?}

\author{
    \parbox{\textwidth}{\centering
        Daniel Panangian\thanks{Corresponding author}\hspace{1cm}
        Ksenia Bittner \\
        The Remote Sensing Technology Institute \\  
        German Aerospace Center (DLR), Wessling, Germany \\
        \tt\small{\{daniel.panangian, ksenia.bittner\}@dlr.de}
    }
}

\maketitle

%

\begin{figure*}[!htp]
    \centering

    \begin{subfigure}[t]{0.49\textwidth}
        \centering
        \begin{subfigure}[t]{0.49\textwidth}
            \centering
            \includegraphics[width=\textwidth]{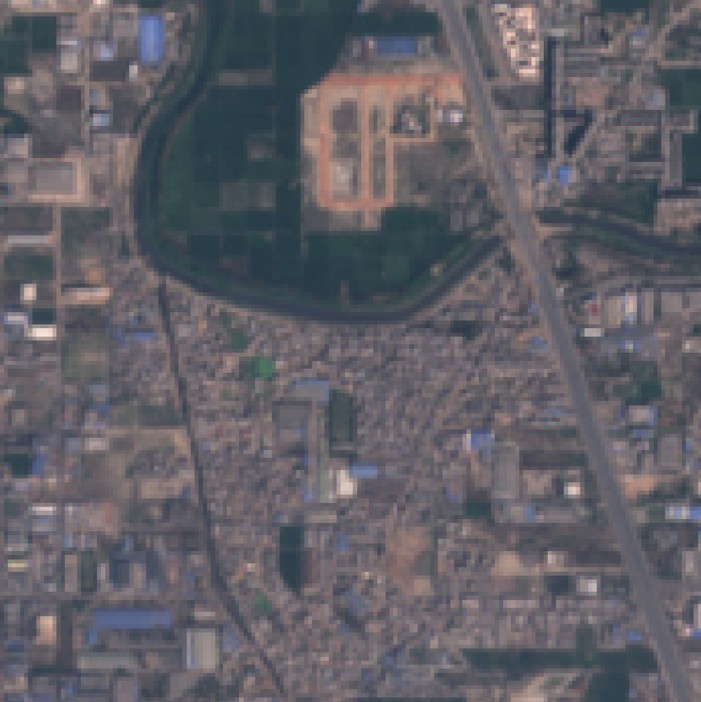} 
        \end{subfigure}
        \hfill
        \begin{subfigure}[t]{0.49\textwidth}
            \centering
            \includegraphics[width=\textwidth]{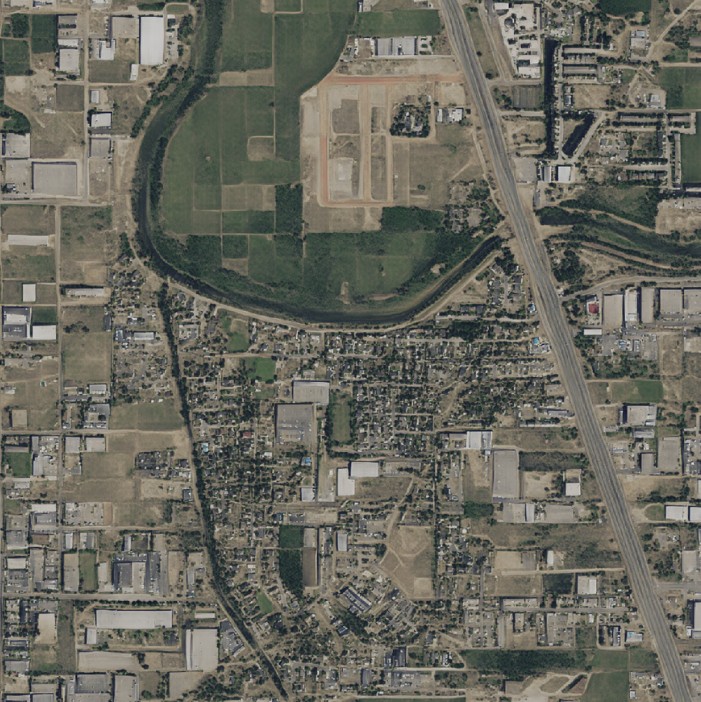} 
        \end{subfigure}
        \caption*{\textbf{Jind, India}}
    \end{subfigure}
    \hfill
    \begin{subfigure}[t]{0.49\textwidth}
        \centering
        \begin{subfigure}[t]{0.49\textwidth}
            \centering
            \includegraphics[width=\textwidth]{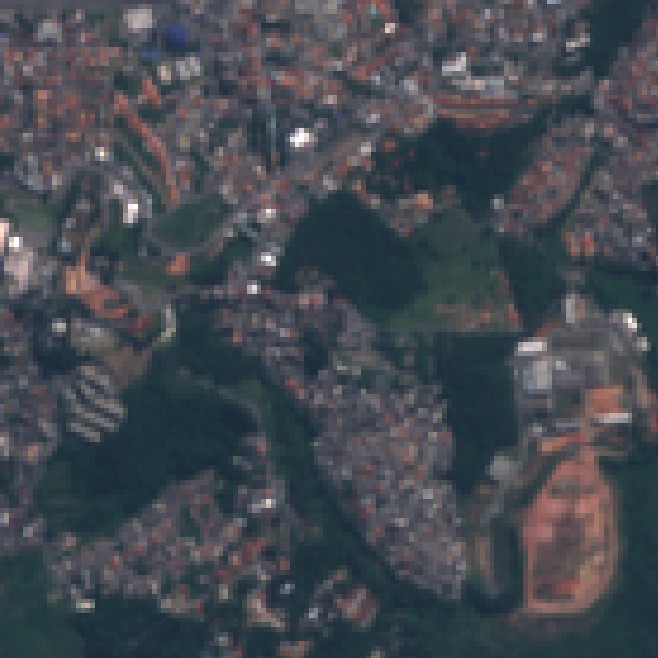} 
        \end{subfigure}
        \hfill
        \begin{subfigure}[t]{0.49\textwidth}
            \centering
            \includegraphics[width=\textwidth]{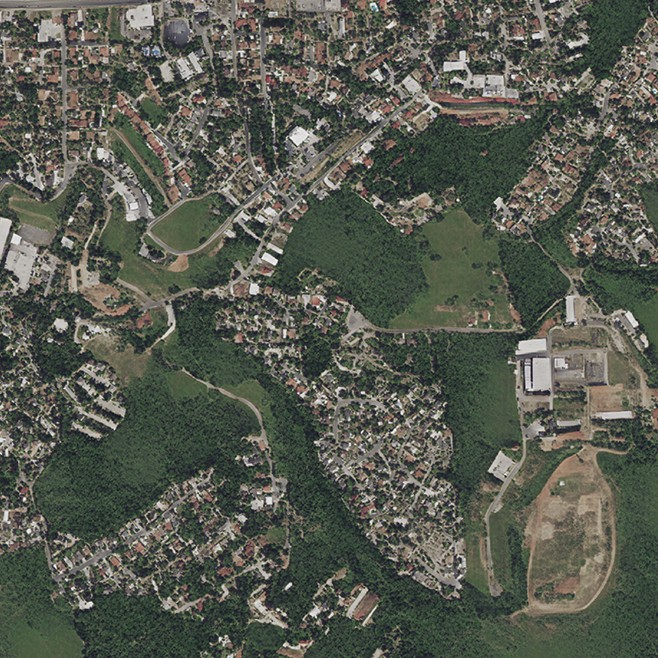} 
        \end{subfigure}
        \caption*{\textbf{São Paulo, Brazil}}
    \end{subfigure}

    \begin{subfigure}[t]{0.49\textwidth}
        \centering
        \begin{subfigure}[t]{0.49\textwidth}
            \centering
            \includegraphics[width=\textwidth]{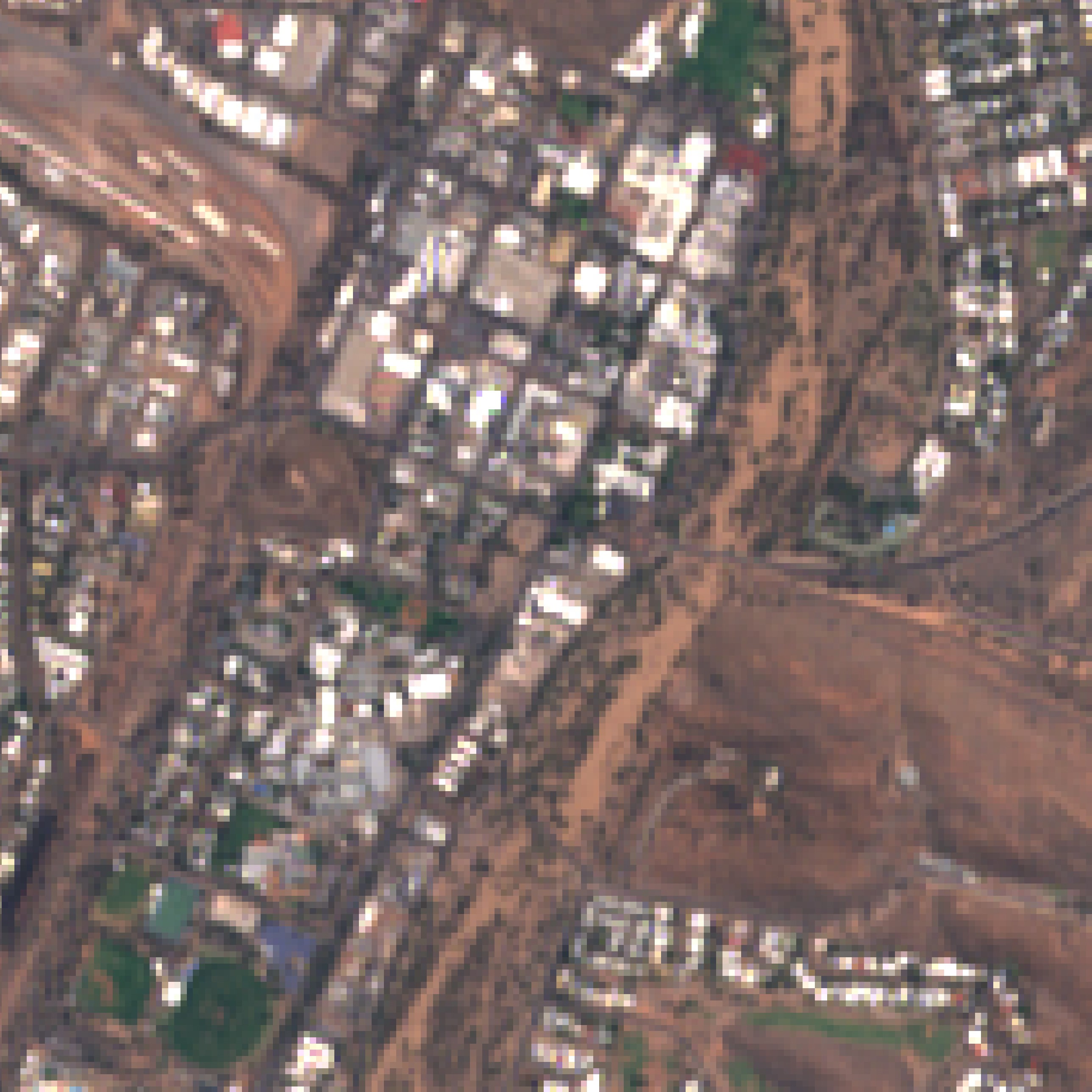} 
        \end{subfigure}
        \hfill
        \begin{subfigure}[t]{0.49\textwidth}
            \centering
            \includegraphics[width=\textwidth]{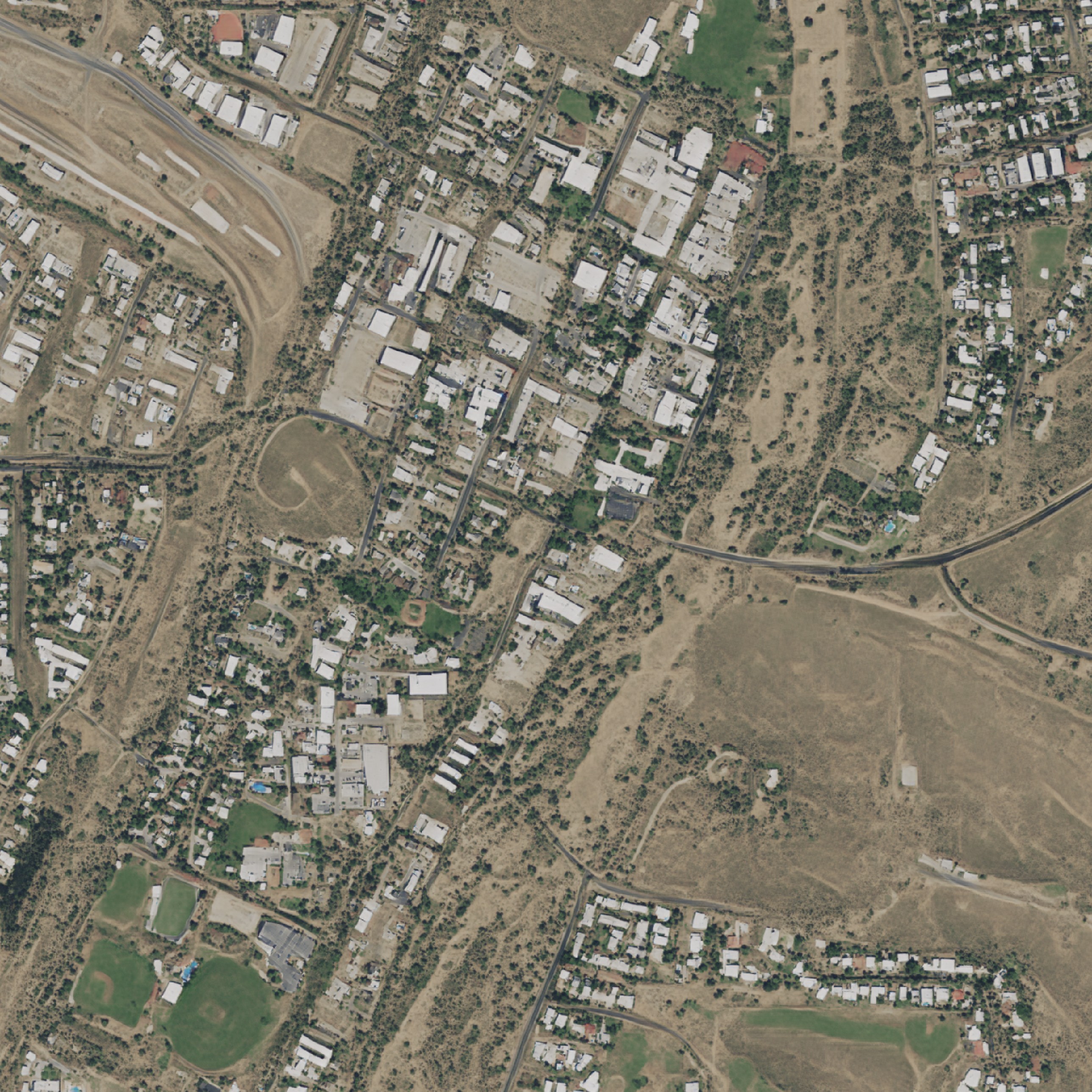} 
        \end{subfigure}
        \caption*{\textbf{Alice Springs, Australia}}
    \end{subfigure}
    \hfill
    \begin{subfigure}[t]{0.49\textwidth}
        \centering
        \begin{subfigure}[t]{0.49\textwidth}
            \centering
            \includegraphics[width=\textwidth]{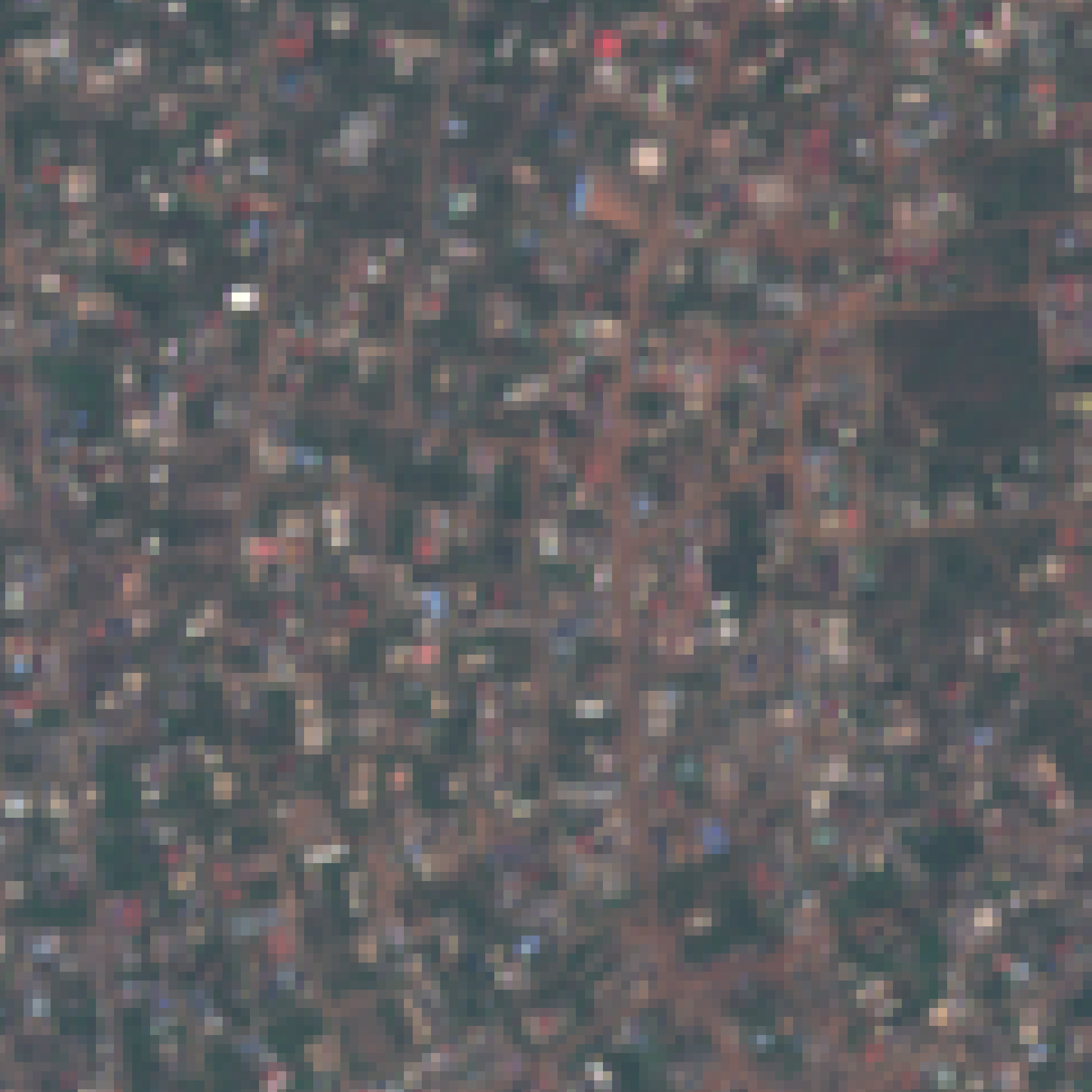} 
        \end{subfigure}
        \hfill
        \begin{subfigure}[t]{0.49\textwidth}
            \centering
            \includegraphics[width=\textwidth]{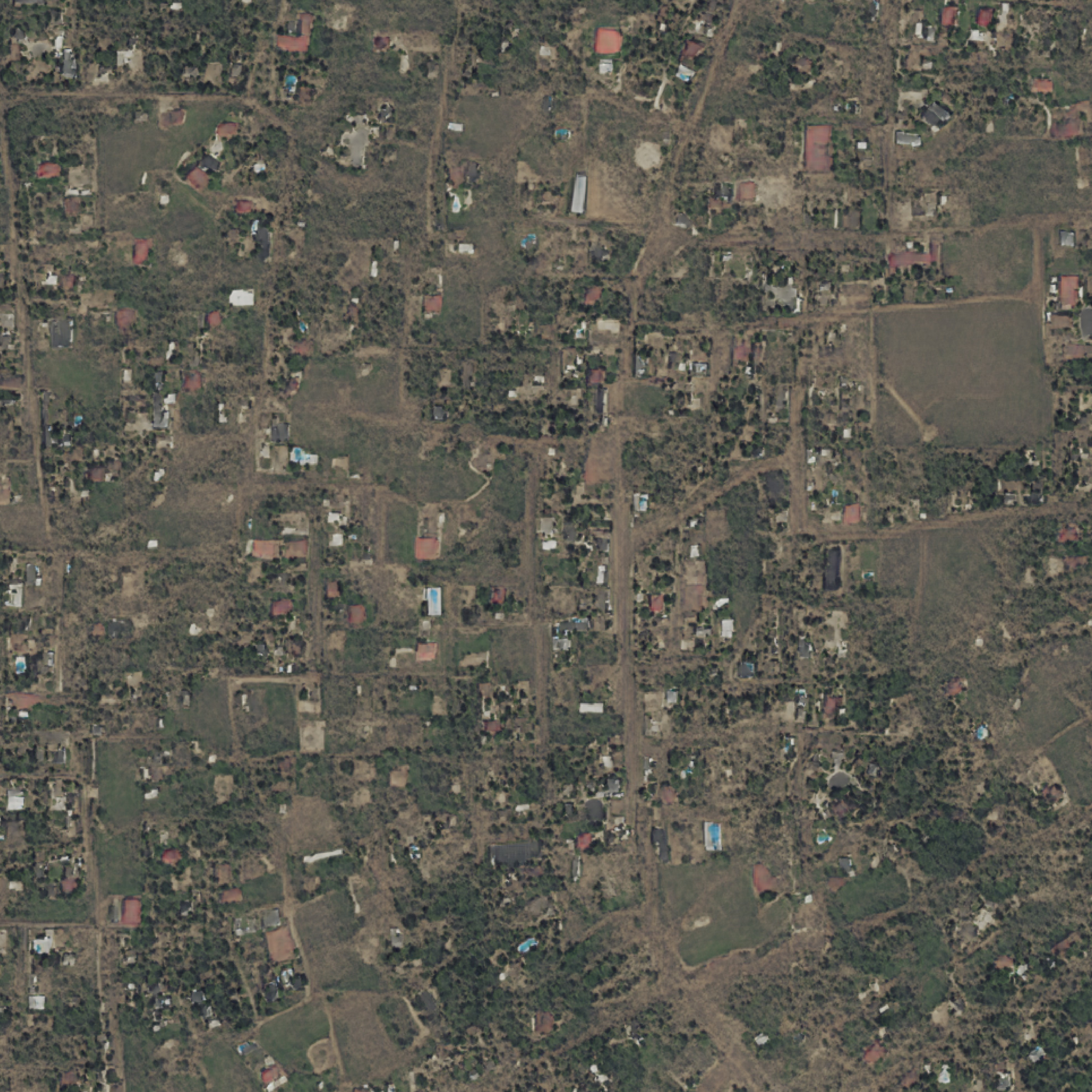} 
        \end{subfigure}
        \caption*{\textbf{Accra, Ghana}}
    \end{subfigure}

    \caption{Visual comparison of satellite images from geographically diverse regions using the proposed super-resolution pipeline. The left image in each pair represents the low-resolution Sentinel-2 input, while the right shows the corresponding super-resolved output, demonstrating the generalization of our method across different global urban landscapes.}
    \label{fig:results}
\end{figure*}

\begin{abstract}
Publicly available satellite imagery, such as Sentinel-2, often lacks the spatial resolution required for accurate analysis of remote sensing tasks including urban planning and disaster response. Current super-resolution techniques are typically trained on limited datasets, leading to poor generalization across diverse geographic regions. In this work, we propose a novel super-resolution framework that enhances generalization by incorporating geographic context through location embeddings. Our framework employs \glspl{gl:GAN} and incorporates techniques from diffusion models to enhance image quality. Furthermore, we address tiling artifacts by integrating information from neighboring images, enabling the generation of seamless, high-resolution outputs. We demonstrate the effectiveness of our method on the building segmentation task, showing significant improvements over state-of-the-art methods and highlighting its potential for real-world applications.
\end{abstract}

\glsresetall

\section{Introduction}

Recent advancements in remote sensing technologies have provided a wealth of satellite imagery, enabling applications in areas such as urban planning, disaster management, environmental monitoring, and resource management. Among these applications, building segmentation is particularly critical, especially for rapidly urbanizing regions. High-resolution imagery is crucial for accurately mapping and analyzing structures; however, limitations in publicly available satellite imagery, such as Sentinel-2, pose challenges due to their relatively low spatial resolution (\SI{10}{}–\SI{20}{\meter} per pixel). To overcome these limitations, super-resolution techniques have emerged as essential tools to enhance image quality and enable finer detail extraction from low resolution images\cite{rs15092347}.

\Glspl{gl:GAN}\cite{goodfellow2014generative} have been previously employed for super-resolution of remote sensing images, but their application has largely been confined to limited datasets, for example NAIP imagery, which covers only regions within the United States \cite{satlassuperres}. Models trained on these datasets often fail to generalize effectively to global regions, producing suboptimal results when applied elsewhere. Moreover when attempting to upscale large areas using tiling techniques, noticeable patch artifacts often emerge, degrading the overall quality of the output\cite{satlassuperres}. 

In recent years, the use of prior information through embeddings has gained popularity, most notably with the success of text embeddings in various tasks. Building on this trend, there has been a growing interest in leveraging geographic context via location embeddings. In this work, we introduce a novel approach that leverages location embedding to enhance super-resolution models. Additionally, we improve the GAN's performance by incorporating techniques used diffusion models. We also conduct experiments to address common patching issues caused by tiling, drawing inspiration from recent advancements in seamless image synthesis.

In summary, our contributions are threefold:
\begin{enumerate}
    \item We develop first location-guided super-resolution model for remote sensing, designed to enhance generalization across diverse geographic regions by integrating location embeddings directly into the model.
    \item We improve a GAN-based super-resolution model's architecture by integrating attention mechanisms techniques to improve the scalability, and context understanding of the model.
    \item We adapt a seamless image synthesis method to super-resolution, tackling common tiling artifacts problem by incorporating neighboring image data, ensuring the generation of continuous, high-resolution satellite imagery.
\end{enumerate}

As shown in \cref{fig:tashkent} and \cref{fig:results}, our results demonstrate super-resolution of low-resolution satellite imagery, marking a key step forward in the application of publicly available dataset for global-scale remote sensing tasks.

\section{Related Works}

\begin{figure*}[!ht]
\begin{center}
\vspace{-0.7cm}
\includegraphics[width=17cm]{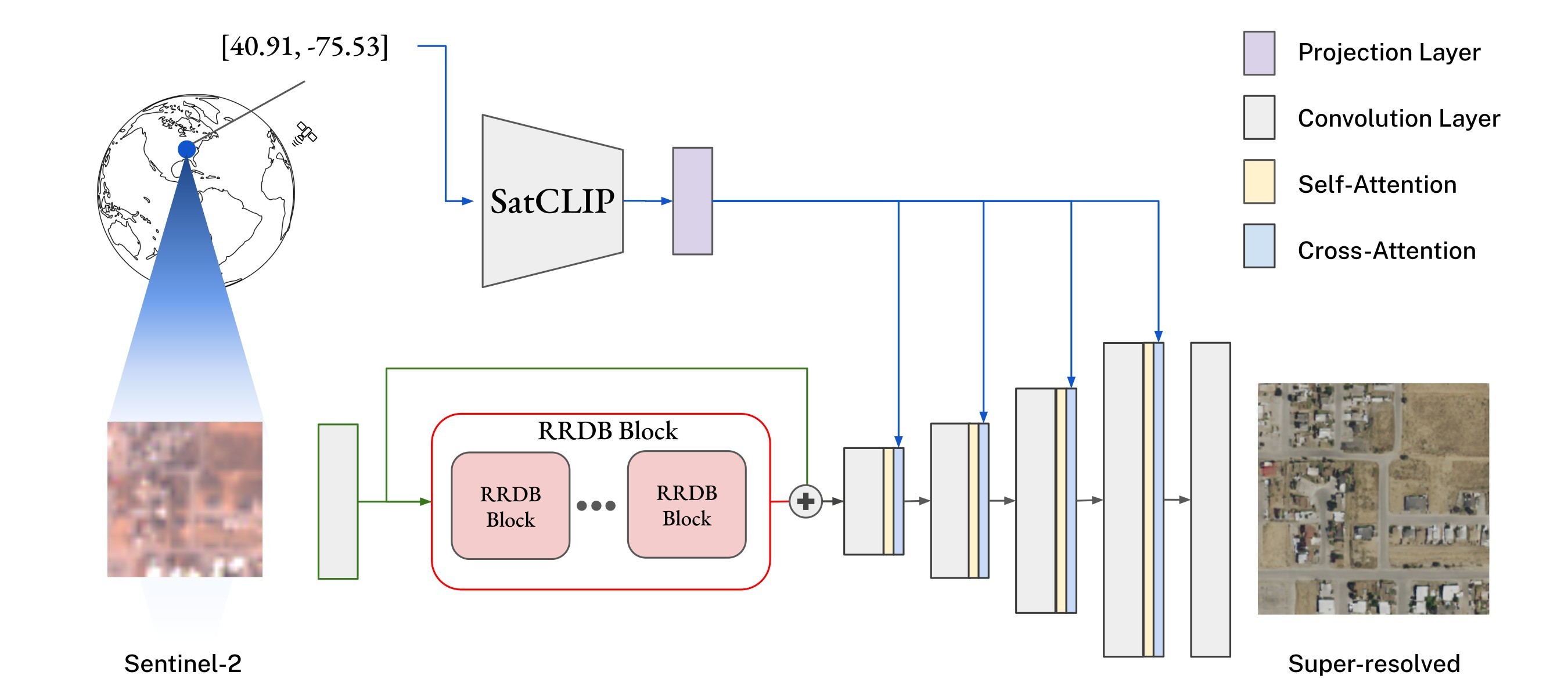}
\end{center}
\vspace{-0.7cm}
   \caption{Summary of the proposed architecture. The input image from Sentinel-2 is processed through a shallow feature extraction network (CNN), followed by the RRDB block for deep feature extraction. Location embeddings, extracted using the SatCLIP module, are projected via MLP layers (purple) and injected into the network during the upsampling phase. The self-attention (yellow) and cross-attention (blue) mechanisms facilitate the integration of the image features and location embeddings, enabling the network to produce a super-resolved image that is both high in visual quality and geographically consistent.}

\label{fig:generator}
\end{figure*}

\subsection{Remote-Sensing Super-Resolution}
Super-resolution techniques in remote sensing have evolved from early \gls{gl:CNN} models like SRCNN \cite{dong2016srcnn} to more advanced generative models. \gls{gl:CNN}-based approaches, such as RCAN \cite{zhang2018rcan}, use attention mechanisms to enhance important image features, resulting in sharper outputs. However, these models are computationally demanding and struggle to scale efficiently for large global datasets. \Glspl{gl:GAN} have further enhanced SR by introducing adversarial loss functions to improve perceptual quality. Models like SRGAN \cite{ledig2017srgan} and ESRGAN \cite{wang2018esrgan} generate realistic textures and high-frequency details with residual-in-residual dense blocks, making them suitable for large-scale remote sensing tasks due to their faster inference times. However, \gls{gl:GAN}-based methods can still introduce artifacts in regions with complex textures, affecting reliability. More recently, diffusion models like SR3 \cite{saharia2021sr3} have become state-of-the-art for generating high-resolution images, utilizing recursive refinement and attention mechanisms to surpass \glspl{gl:CNN} and \glspl{gl:GAN} in visual fidelity. However, their high computational demands and slow processing make them less practical for large-scale satellite imagery applications. At the same time, there's a growing trend of integrating priors or different modalities to control and improve image generation and restoration. Models like GLEAN~\cite{chan2021glean} and FeMaSR~\cite{femasr} utilize pretrained large generative models, such as StyleGAN\cite{Karras2019stylegan2} or VQGAN\cite{esser2020taming} for super-resolution. Some approaches also use modalities like text prompts to guide image generation \cite{rombach2021high,Gandikota_2024_CVPR}. In the remote sensing domain, methods like DSen2~\cite{lanaras} and Real-GDSR~\cite{panangian2024real} use multi-modal inputs, while DiffusionSAT \cite{khanna2024diffusionsat} integrates metadata, such as time and geographic coordinates, to control image generation via conditional generative trained on big large data\cite{rombach2021high}. Diffusion-based techniques offer greater control and flexibility but their scalability remains a significant challenge, especially for large-scale remote sensing applications.

The datasets available for remote sensing super-resolution largely dictate the performance and generalization of the models trained on them. Many datasets use low-resolution Sentinel-2 imagery due to its free availability and balanced temporal and spatial resolution \cite{sen2venms,worldstrat}. High-resolution data is often sourced from Google Earth \cite{Dong2021RRSGANRS}, PlanetScope\cite{Razzak2021MultiSpectralMS}, or WorldView \cite{article}, which are regionally constrained and fail to cover diverse geographic conditions, which limits the practical applicability of models trained on them. To address this, we are exploring the use of available modalities and features to improve model generalization without relying solely on expensive high-resolution data. 

\subsection{Location Encoders}

Location encoders transform geographic coordinates into rich, contextual embeddings that capture environmental and spatial dependencies, such as terrain, climate, and human land use. These models turned out to be effective in tasks like geolocation, spatial prediction, and environmental monitoring by embedding geographic context into deep learning systems. Models like GPS2Vec\cite{yin2019gps2vec} and CSP\cite{mai2023contrastive} have applied location-based embeddings to tasks such as image retrieval and species classification, but they faced limitations in generalization due to uneven data distribution across geographic regions. GeoCLIP\cite{cepeda2023geoclip} further refined geolocation tasks using contrastive learning but still encountered similar biases. A more comprehensive approach emerged with SatCLIP\cite{klemmer2023satclip}, which uses satellite data to generate embeddings that generalize across global landscapes, allowing it to perform well in tasks like population density estimation and environmental prediction.

While recent developments like GeoSynth~\cite{Sastry_2024_CVPR} have used these embeddings for satellite image generation, our work focuses on using geographic location encoders for image super-resolution. By incorporating location-based features from models like SatCLIP, we aim to enhance satellite image resolution while accounting for diverse geographic and environmental conditions. This approach allows our super-resolution models to perform more effectively in remote sensing tasks by embedding spatial context directly into the resolution enhancement process, improving image clarity across varying terrains and climates.

\section{Method}
\subsection{Problem Formulation}
Given a low-resolution satellite image and its corresponding geographic location, our goal is to reconstruct a high-resolution image that reflects both fine visual details and the correct geographic characteristics. Unlike traditional super-resolution, where reconstruction relies solely on pixel data, our approach integrates location-based context to ensure the image aligns with real-world geography.

To achieve this, we address two main questions:
\begin{itemize}
        \item \textbf{How can we integrate location embeddings into the super-resolution network?} These embeddings, can be integrated by concatenating or modulating them with image feature maps at various stages of the network. This allows the network to incorporate spatial context, enhancing the super-resolved image’s geographic and environmental consistency.

    \item \textbf{How do we design better conditioning to control the generated image?} The location data must be integrated effectively with image features to ensure the reconstructed image is geographically consistent.
\end{itemize}

 We introduce a cross-attention module to fuse the features from the low-resolution image with the location embeddings. The cross-attention mechanism allows the model to dynamically focus on regions of the image that require more refinement based on the geographic context. This helps the network associate specific textures or colors, such as vegetation or urban areas, with the location, enhancing the image's overall realism and detail. To further enhance the geographic consistency of the generated high-resolution images, we design a location-matching discriminator. This discriminator ensures that the textures and visual patterns in the high-resolution image align with the geographic context provided by the location embeddings.
\subsection{Model Overview}

As illustrated in~\cref{fig:generator}, our proposed super-resolution model integrates satellite imagery and location embeddings for high-resolution image reconstruction. The architecture begins with the extraction of shallow features via a convolutional layer, which capture basic image attributes like edges and textures. The \glspl{gl:RRDB} then processes these shallow features to capture high-level, complex details essential for high-fidelity super-resolution.

A key element of our model is the integration of SatCLIP embeddings, which capture geographical context and urban patterns based on the image's coordinates. This step plays a vital role in incorporating location-specific information, helping the model generalize effectively across diverse regions. The SatCLIP embeddings are projected using \glspl{gl:MLP} and are introduced into the network during the upsampling phase. We chose this, as the upsampling stage focuses on refining elements like texture and color without modifying the spatial features that were already captured during earlier feature extraction. By injecting the embeddings at this phase, the location-specific details such as textures and colors are adjusted in a way that enhances realism, while preserving the spatial structure.
\subsection{Super-Resolution Network}
Our model builds upon the foundation of ESRGAN\cite{wang2018esrgan}, a state-of-the-art framework for image super-resolution. ESRGAN improves over traditional SRGAN by using \gls{gl:RRDB} blocks, which enhance the network’s capacity and ease of training without the need for batch normalization. These blocks are designed to improve the recovery of complex textures and reduce artifacts, a challenge seen in traditional SRGAN. In this work we try to further improve it by utilizing self-attention mechanisms in the upsampling step to allow the network to focus on different parts of the image, capturing long-range dependencies and spatial relationships between regions. Applying self-attention at each stage helps the model maintain spatial consistency across different resolutions and overall coherence during the super-resolution process.

\begin{figure}[t] 
    \begin{center} 
    \includegraphics[width=0.45\textwidth]{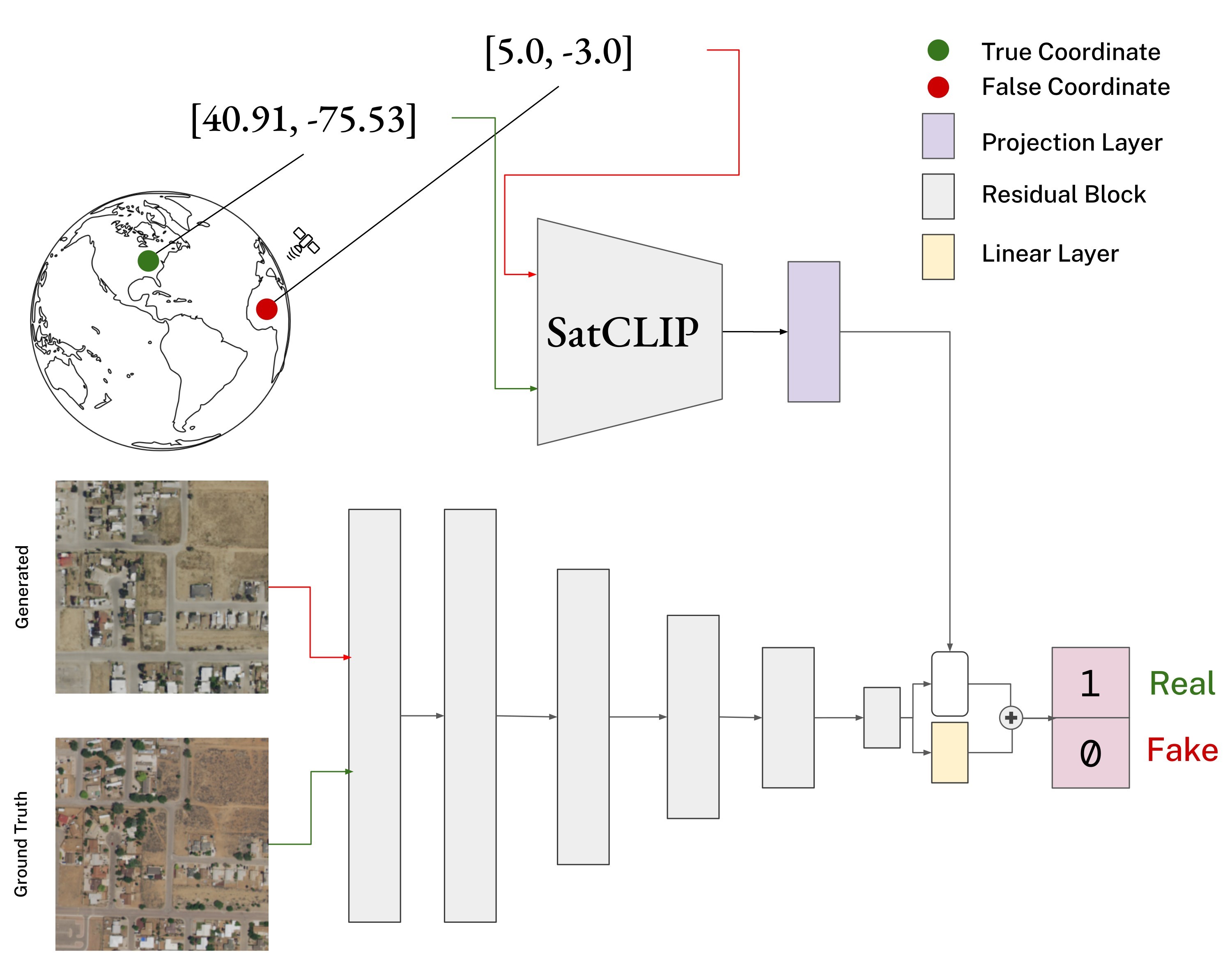} 
     \put (-53,50){\rotatebox[origin=c]{-90}{\scriptsize{a$\cdot$b}}}
     \put (-54,38){\scriptsize{$\psi$}}
    \end{center} 
    \caption{Overview of the location-matching Discriminator. The input consists of an image paired with real and false geographic coordinates, processed through the SatCLIP module. The image and the location embeddings are then passed through the discriminator network. The output of the discriminator is a binary classification (real or fake) based on whether the location matches the input image. The SatCLIP location embeddings guide the discriminator, enforcing geographic consistency in the image generation process.}
    \label{fig:discriminator}
\end{figure}

\subsection{Location Conditioning}

We then introduce location conditioning through cross-attention after self-attention mechanisms. With the image features serving as the query, and the location embeddings as key-value pairs. This enables the model to fuse visual data from the image with the corresponding geographic context. This approach allows the model to first establish the internal consistency before applying more complex, external guidance in each upsampling stage. 

However, despite these advantages, we add more explicit mechanism to evaluate and guide the generation process. We introduce a location-matching discriminator into the model. This discriminator not only evaluates the visual similarity between super-resolved and real images but also ensures geographic accuracy by comparing generated images with their true locations. Additionally, during training, we introduce random false locations, forcing the discriminator to distinguish between correct and incorrect geographic contexts. This encourages the model to not only integrate location information effectively but also ensures that the generated image consistently aligns with their respective location embeddings. As seen in Figure \ref{fig:discriminator} the architecture of the discriminator is inspired by projection discriminator\cite{Miyato2018cGANsWP}, a method designed to incorporate conditional information (such as location embeddings in our case) into the discriminator in a more effective way than traditional methods like concatenation. The projection discriminator applies a projection layer that takes the inner product between the condition and the intermediate features of the image within the discriminator. This leads to more robust conditioning than concatenation, as it directly influences the discriminator's decision-making by embedding the condition deeper into its structure.

\subsection{Training Constraints}

We define the \textit{location-matching loss} as:
\begin{align}
\mathcal{L}_{\text{loc\_match}} = 
&\; \mathbb{E}_{x_{\text{HR}}, x_{\text{LR}}, c, \hat{c}} \Big[ 
  \log D(x_{\text{HR}}, c) \notag \\
& + \log \big(1 - D(x_{\text{HR}}, \hat{c})\big) \notag \\
& + \log \big(1 - D(G(x_{\text{LR}}, c), \hat{c})\big)
\Big]
\end{align}
where \( x_{\text{HR}} \) represents the high-resolution image, 
\( x_{\text{LR}} \) is the corresponding low-resolution image, 
\( c \) denotes the true location embedding, and 
\( \hat{c} \) is a false location embedding. 
This loss encourages the model to correctly associate geographic information by ensuring that images are matched with their true locations while distinguishing them from false ones.

The total objective function is:
\begin{equation}
\begin{aligned}
L(G, D) = &\ L_{\text{pix}} + L_{\text{perceptual}} + L_{\text{CLIP}} \\
          &+ L_{\text{OSM}} + L_{\text{loc\_match}}
\end{aligned}
\end{equation}

While pixel loss and perceptual loss handle pixel-level and perceptual fidelity, we also incorporate CLIP loss and OpenStreetMap (OSM) discriminator loss from Satlas-SR~\cite{satlassuperres}. CLIP loss ensures that the generated images align with high-level semantic content, while OSM loss ensures accurate reconstruction of geographic entities like buildings and roads based on OSM data. We enhance the OSM loss by preserving spatial relationships between these objects, ensuring geographic coherence in the generated images. Together, these losses balance visual quality and geographic accuracy. Our addition of the location-matching loss further enforces spatial correctness, ensuring that the generated images are not only perceptually accurate but also location-guided.

\subsection{Local Padding}
At inference, we replace all padding with adjacent neighboring patches, following the technique proposed by Latif et al. \cite{abdellatif2024localpaddingpatchbasedgans}. Instead of using conventional zero-padding, which can introduce artificial boundaries, this approach leverages the information from surrounding patches. By incorporating neighboring patches into the padding, we ensure smoother transitions and eliminate visible seams that can occur when concatenating patches. This method results in more consistent and seamless outputs during inference.



\section{Experiments}

\subsection{Dataset}
We utilize the S2-NAIP dataset \cite{satlassuperres}, which pairs low-resolution Sentinel-2 images (64×64 pixels) with high-resolution National Agriculture Imagery Program (NAIP) imagery (512×512 pixels). This dataset is geographically situated within the United States. We apply random cropping to the Sentinel-2 images, reducing them to 32×32 pixels, while concurrently cropping the corresponding NAIP images to 256×256 pixels.

\begin{figure}[t]
\begin{center}
\includegraphics[width=0.48\textwidth]{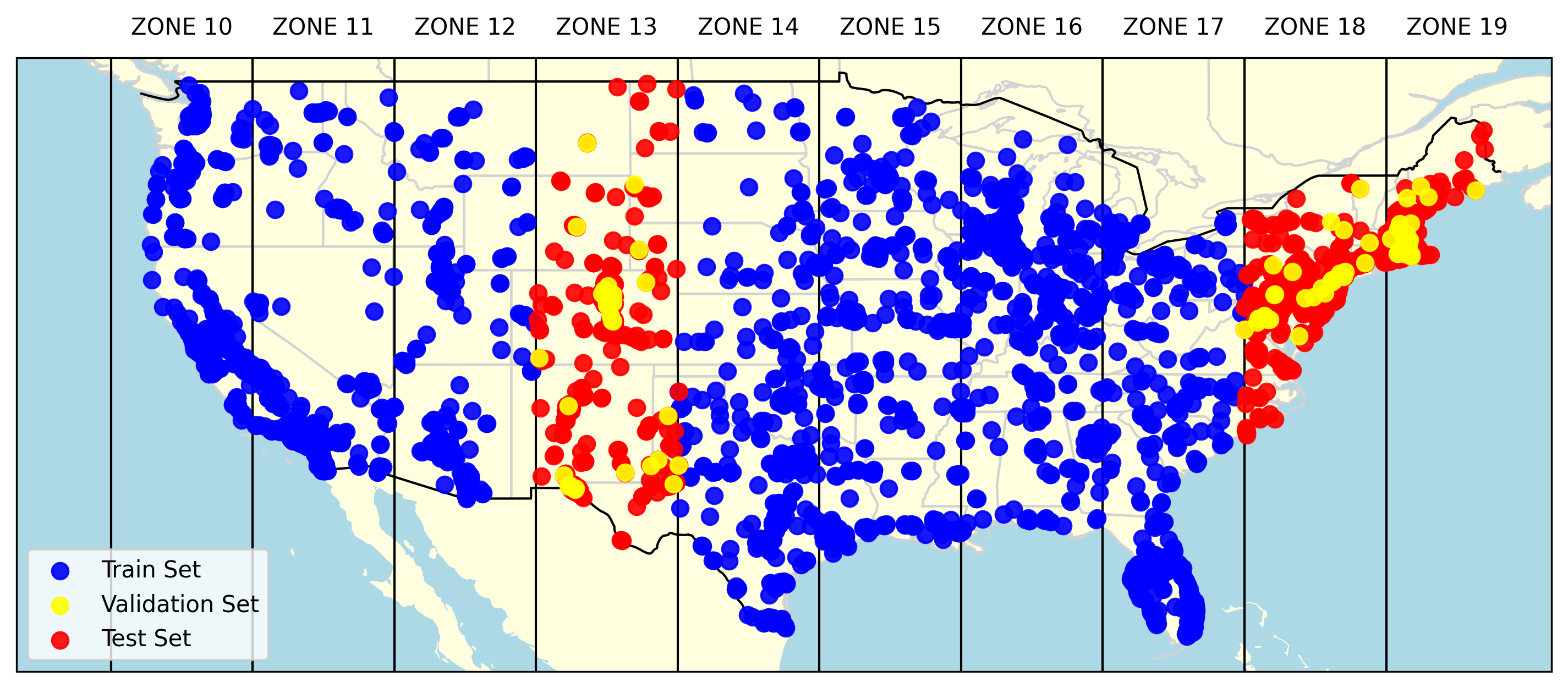}
\end{center}
   \caption{Visualization of the data split by UTM zones. The train set includes samples from UTM zones 10-17, while the validation and test set consists of samples from UTM zones 13N, 18N, and 19N. These zones were selected to evaluate how well the model generalizes across different geographic regions.}
\label{fig:utm_zones}
\end{figure}

The dataset is geographically divided by \gls{gl:UTM} zones. We chose \gls{gl:UTM} zones 13N, 18N, and 19N as the validation and test set for our experiments, as illustrated in Figure \ref{fig:utm_zones}. These zones were selected based on their spatial positioning relative to the training zones: zone 13N is located in the middle of the training zones, while 18N and 19N are located at the far east. This geographic separation allows us to rigorously test the generalization ability of the model across different landscapes. We select central and edge UTM zones to assess the efficacy of embedding mechanisms within our model. Central UTM zones represent regions centrally located within the dataset's spatial distribution, providing a baseline for typical geographic variability. In contrast, edge UTM zones are chosen from the peripheries to introduce challenges related to spatial extrapolation and embedding generalization.

\subsection{Implementation Details}

Our model is implemented using the BasicSR\cite{basicsr} framework, a widely-used platform for super-resolution tasks. To benchmark our approach, we compare it against Satlas-SR\cite{satlassuperres}, which serves as our baseline. Satlas-SR primarily contributes by integrating CLIP loss for faster convergence \cite{radford2021clip} and an object-aware discriminator that ensures spatial consistency. These components are incorporated into our evaluation to provide a fair and comprehensive comparison.
Additionally, we compare our results with Stable Diffusion \cite{rombach2021high} with ControlNet\cite{zhang2023adding}, which focuses on conditional image generation rather than traditional super-resolution. ControlNet enhances Stable Diffusion by allowing the network to condition on additional control signals, such as edge maps or segmentation maps, to generate images that adhere more strictly to the provided conditions. A similar approach is explored in DiffusionSAT \cite{khanna2024diffusionsat}, which uses diffusion models for remote-sensing image generation and other tasks such as SR and inpainting by leveraging metadata like geolocation and temporal information. Due to the unavailability of its conditional generation code at the time of our experiments, we did not include DiffusionSAT in our evaluation.

For quantitative evaluation, we use four key metrics: Peak Signal-to-Noise Ratio (PSNR) and Structural Similarity Index Measure (SSIM) for pixel-level accuracy, Learned Perceptual Image Patch Similarity (LPIPS) \cite{zhang2018lpips} for perceptual quality, and the CLIP score \cite{radford2021clip} to evaluate image-text alignment. In addition to these metrics, we evaluate the performance of the super-resolved images on semantic segmentation task. Specifically, we fine-tune the Segment Anything Model (SAM)\cite{kirillov2023segany} on NAIP imagery and building segmentation labels derived from OSM. We then assess the quality of the segmentation using metrics such as mean Intersection over Union (mIoU), F1 score for building detection, and mean F1 score (mF1) for buildings. This provides additional insights into the effectiveness of our super-resolution model in solving downstream tasks in remote sensing. More detailed descriptions of baselines, hyperparameters, and training details can be found in the supplementary Sec. \textcolor{red}{S2}.

\begin{figure*}[htbp]
    \centering

    \begin{subfigure}[b]{0.19\textwidth}
        \centering
        \begin{tikzpicture}
            \node[anchor=south west, inner sep=0] at (0,0) {\includegraphics[height=3.3cm]{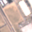}};
            \draw[red, thick] (1.8,1.8) rectangle ++(1,1); 
        \end{tikzpicture}
    \end{subfigure}
    \begin{subfigure}[b]{0.19\textwidth}
        \centering
        \begin{tikzpicture}
            \node[anchor=south west, inner sep=0] at (0,0) {\includegraphics[height=3.3cm]{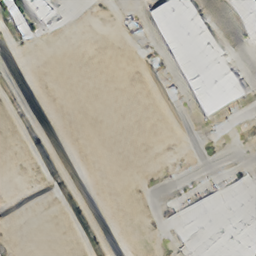}};
           \draw[red, thick] (1.8,1.8) rectangle ++(1,1);
        \end{tikzpicture}
    \end{subfigure}
    \begin{subfigure}[b]{0.19\textwidth}
        \centering
        \begin{tikzpicture}
            \node[anchor=south west, inner sep=0] at (0,0) {\includegraphics[height=3.3cm]{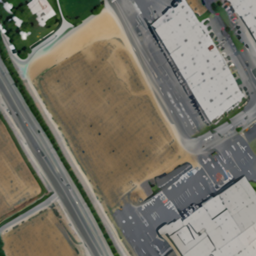}};
            \draw[red, thick] (1.8,1.8) rectangle ++(1,1);
        \end{tikzpicture}
    \end{subfigure}
    \begin{subfigure}[b]{0.19\textwidth}
        \centering
        \begin{tikzpicture}
            \node[anchor=south west, inner sep=0] at (0,0) {\includegraphics[height=3.3cm]{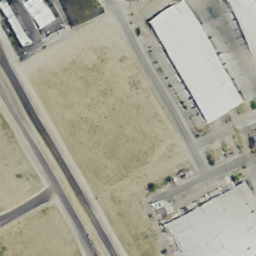}};
            \draw[red, thick] (1.8,1.8) rectangle ++(1,1);
        \end{tikzpicture}
    \end{subfigure}
    \begin{subfigure}[b]{0.19\textwidth}
        \centering
        \begin{tikzpicture}
            \node[anchor=south west, inner sep=0] at (0,0) {\includegraphics[height=3.3cm]{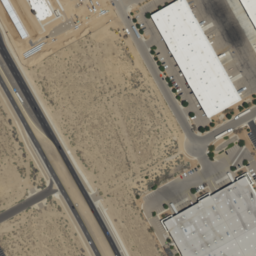}};
            \draw[red, thick] (1.8,1.8) rectangle ++(1,1);
        \end{tikzpicture}
    \end{subfigure}

    \vspace{1em} 

    \begin{subfigure}[b]{0.19\textwidth}
        \centering
        \begin{tikzpicture}
            \node[anchor=south west, inner sep=0] at (0,0) {\includegraphics[height=3.3cm]{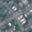}};
            \draw[red, thick] (1.8,2.5) rectangle ++(0.7,0.7);
        \end{tikzpicture}
    \end{subfigure}
    \begin{subfigure}[b]{0.19\textwidth}
        \centering
        \begin{tikzpicture}
            \node[anchor=south west, inner sep=0] at (0,0) {\includegraphics[height=3.3cm]{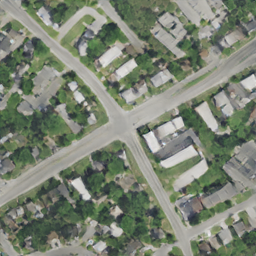}};
            \draw[red, thick] (1.8,2.5) rectangle ++(0.7,0.7);
        \end{tikzpicture}
    \end{subfigure}
    \begin{subfigure}[b]{0.19\textwidth}
        \centering
        \begin{tikzpicture}
            \node[anchor=south west, inner sep=0] at (0,0) {\includegraphics[height=3.3cm]{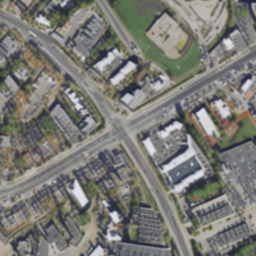}};
            \draw[red, thick] (1.8,2.5) rectangle ++(0.7,0.7);
        \end{tikzpicture}
    \end{subfigure}
    \begin{subfigure}[b]{0.19\textwidth}
        \centering
        \begin{tikzpicture}
            \node[anchor=south west, inner sep=0] at (0,0) {\includegraphics[height=3.3cm]{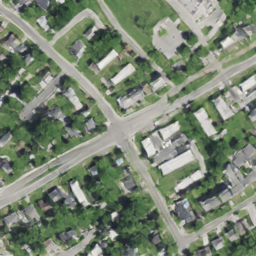}};
            \draw[red, thick] (1.8,2.5) rectangle ++(0.7,0.7);
        \end{tikzpicture}
    \end{subfigure}
    \begin{subfigure}[b]{0.19\textwidth}
        \centering
        \begin{tikzpicture}
            \node[anchor=south west, inner sep=0] at (0,0) {\includegraphics[height=3.3cm]{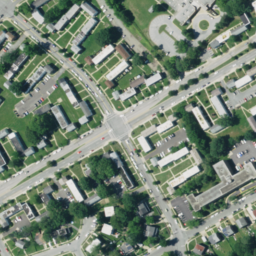}};
            \draw[red, thick] (1.8,2.5) rectangle ++(0.7,0.7);
        \end{tikzpicture}
    \end{subfigure}

    \vspace{1em} 

    \begin{subfigure}[b]{0.19\textwidth}
        \centering
        \begin{tikzpicture}
            \node[anchor=south west, inner sep=0] at (0,0) {\includegraphics[height=3.3cm]{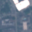}};
            \draw[red, thick] (0.1,1.5) rectangle ++(1.4,0.5);
        \end{tikzpicture}
        \caption{Sentinel-2}
    \end{subfigure}
    \begin{subfigure}[b]{0.19\textwidth}
        \centering
        \begin{tikzpicture}
            \node[anchor=south west, inner sep=0] at (0,0) {\includegraphics[height=3.3cm]{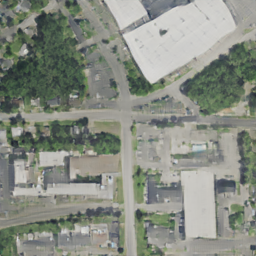}};
            \draw[red, thick] (0.1,1.5) rectangle ++(1.4,0.5);
        \end{tikzpicture}
        \caption{Satlas-SR}
    \end{subfigure}
    \begin{subfigure}[b]{0.19\textwidth}
        \centering
        \begin{tikzpicture}
            \node[anchor=south west, inner sep=0] at (0,0) {\includegraphics[height=3.3cm]{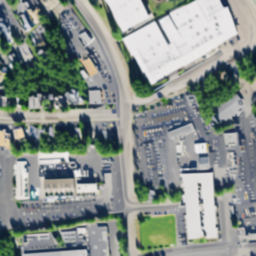}};
            \draw[red, thick] (0.1,1.5) rectangle ++(1.4,0.5);
        \end{tikzpicture}
        \caption{SD + ControlNet}
    \end{subfigure}
    \begin{subfigure}[b]{0.19\textwidth}
        \centering
        \begin{tikzpicture}
            \node[anchor=south west, inner sep=0] at (0,0) {\includegraphics[height=3.3cm]{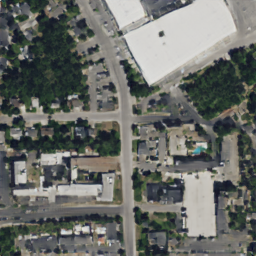}};
            \draw[red, thick] (0.1,1.5) rectangle ++(1.4,0.5);
        \end{tikzpicture}
        \caption{Ours}
    \end{subfigure}
    \begin{subfigure}[b]{0.19\textwidth}
        \centering
        \begin{tikzpicture}
            \node[anchor=south west, inner sep=0] at (0,0) {\includegraphics[height=3.3cm]{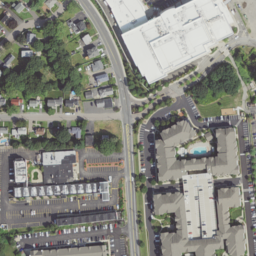}};
            \draw[red, thick] (0.1,1.5) rectangle ++(1.4,0.5);
        \end{tikzpicture}
        \caption{NAIP}
    \end{subfigure}

        \caption{Performance comparison of super-resolution methods on satellite imagery from the test set. Each row shows a distinct scene with the Sentinel-2 input, super-resolved outputs, and NAIP image for reference.}
    \label{fig:super-resolution-metrics}
\end{figure*}

\begin{figure*}[htbp]
    \centering

    \begin{subfigure}[b]{0.19\textwidth}
        \centering
        \begin{tikzpicture}
            \node[anchor=south west, inner sep=0] at (0,0) {\includegraphics[height=3.3cm]{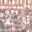}};
            \draw[red, thick] (0.1,2.3) rectangle ++(2.2,0.5);
            \draw[red, thick] (2.7,0.2) rectangle ++(0.5,2.2);
        \end{tikzpicture}
    \end{subfigure}
    \hfill
    \begin{subfigure}[b]{0.19\textwidth}
        \centering
        \begin{tikzpicture}
            \node[anchor=south west, inner sep=0] at (0,0) {\includegraphics[height=3.3cm]{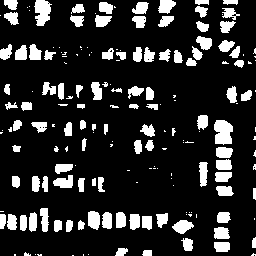}};
            \draw[red, thick] (0.1,2.3) rectangle ++(2.2,0.5);
            \draw[red, thick] (2.7,0.2) rectangle ++(0.5,2.2);
        \end{tikzpicture}
    \end{subfigure}
    \hfill
    \begin{subfigure}[b]{0.19\textwidth}
        \centering
        \begin{tikzpicture}
            \node[anchor=south west, inner sep=0] at (0,0) {\includegraphics[height=3.3cm]{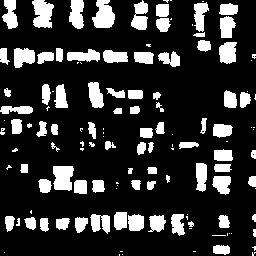}};
            \draw[red, thick] (0.1,2.3) rectangle ++(2.2,0.5);
            \draw[red, thick] (2.7,0.2) rectangle ++(0.5,2.2);
        \end{tikzpicture}
    \end{subfigure}
    \hfill
    \begin{subfigure}[b]{0.19\textwidth}
        \centering
        \begin{tikzpicture}
            \node[anchor=south west, inner sep=0] at (0,0) {\includegraphics[height=3.3cm]{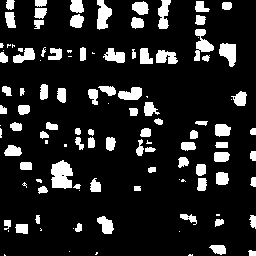}};
            \draw[red, thick] (0.1,2.3) rectangle ++(2.2,0.5);
            \draw[red, thick] (2.7,0.2) rectangle ++(0.5,2.2);
        \end{tikzpicture}
    \end{subfigure}
    \hfill
    \begin{subfigure}[b]{0.19\textwidth}
        \centering
        \begin{tikzpicture}
            \node[anchor=south west, inner sep=0] at (0,0) {\includegraphics[height=3.3cm]{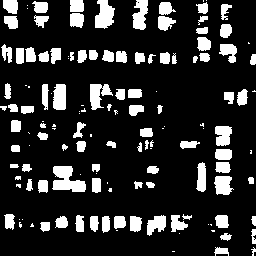}};
            \draw[red, thick] (0.1,2.3) rectangle ++(2.2,0.5);
            \draw[red, thick] (2.7,0.2) rectangle ++(0.5,2.2);
        \end{tikzpicture}
    \end{subfigure}
    
    \vspace{1em} 
    
    \begin{subfigure}[b]{0.19\textwidth}
        \centering
        \begin{tikzpicture}
            \node[anchor=south west, inner sep=0] at (0,0) {\includegraphics[height=3.3cm]{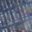}};
            \draw[red, thick] (2,0.9) rectangle ++(0.8,2);
        \end{tikzpicture}
        \caption{Sentinel-2}
    \end{subfigure}
    \hfill
    \begin{subfigure}[b]{0.19\textwidth}
        \centering
        \begin{tikzpicture}
            \node[anchor=south west, inner sep=0] at (0,0) {\includegraphics[height=3.3cm]{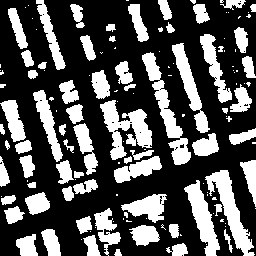}};
            \draw[red, thick] (2,0.9) rectangle ++(0.8,2);
        \end{tikzpicture}
        \caption{NAIP}
    \end{subfigure}
    \hfill
    \begin{subfigure}[b]{0.19\textwidth}
        \centering
        \begin{tikzpicture}
            \node[anchor=south west, inner sep=0] at (0,0) {\includegraphics[height=3.3cm]{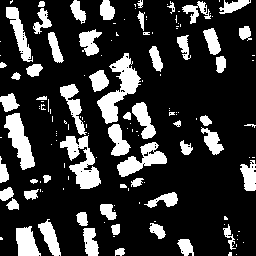}};
            \draw[red, thick] (2,0.9) rectangle ++(0.8,2);
        \end{tikzpicture}
        \caption{Satlas-SR}
    \end{subfigure}
    \hfill
    \begin{subfigure}[b]{0.19\textwidth}
        \centering
        \begin{tikzpicture}
            \node[anchor=south west, inner sep=0] at (0,0) {\includegraphics[height=3.3cm]{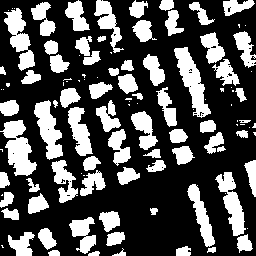}};
            \draw[red, thick] (2,0.9) rectangle ++(0.8,2);
        \end{tikzpicture}
        \caption{SD + ControlNet}
    \end{subfigure}
    \hfill
    \begin{subfigure}[b]{0.19\textwidth}
        \centering
        \begin{tikzpicture}
            \node[anchor=south west, inner sep=0] at (0,0) {\includegraphics[height=3.3cm]{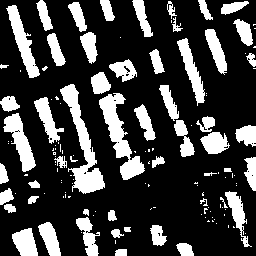}};
            \draw[red, thick] (2,0.9) rectangle ++(0.8,2);
        \end{tikzpicture}
        \caption{Ours}
    \end{subfigure}
    
    \caption{Building segmentation results of different super-resolution methods on the test set. The NAIP column represents the results of the segmentation model using NAIP images as input, serving as the reference for evaluating the accuracy of the other approaches.}
    \label{fig:segmentation-metrics}
\end{figure*}

\section{Results}
In this section, we present both qualitative and quantitative comparisons of our model against the baselines, including Satlas-SR and Stable Diffusion with ControlNet (SD + ControlNet). We evaluate the models on super-resolution and building segmentation tasks, showcasing the effectiveness of our approach. Additionally, we provide results on the impact of local padding in the supplementary Sec. \textcolor{red}{S3}.

\subsection{Super-Resolution}

In \cref{tab:super-resolution-metrics}, we observe that Satlas-SR delivers the highest PSNR and LPIPS value and of 15.6481, indicating superior pixel-level reconstruction accuracy. However, our model slightly outperforms all others in CLIP score, achieving 0.9261, which highlights its strong alignment with reference data, particularly in terms of semantic relevance. This demonstrates competitive performance of our model in both pixel-level fidelity and perceptual accuracy with other methods.

\Cref{fig:super-resolution-metrics} visually compares the super-resolved images generated by our method and baselines. In the first row, depicting an industrial area, the super-resolved image produced by our model preserves not only finer building edges and textures but also better captures the road structures compared to Satlas-SR and SD + ControlNet. Satlas-SR fails to produce clear outlines and shapes particularly for the buildings and roads. SD + ControlNet, on the other hand, hallucinates vegetation in areas where there is none, leading to visually unrealistic elements in the scene. Our model achieves a better balance between sharpness and naturalness, recovering both roads and building structures better than other baselines.

In the second row, depicting a suburban area, our method offers better contrast and more details, making it more visually aligned with the ground truth. The roads and building edges are more clearly defined, enhancing the overall realism. In comparison, SD + ControlNet does a reasonable job at reconstructing some buildings but introduces repetitive patterns that give the image an unnatural, synthetic appearance. Additionally, the variation in vegetation color creates an inconsistency, making it seem like the image was captured in a different season. Satlas-SR struggles with over-reconstructing buildings, adding structures where there should be none, leading to a cluttered and less accurate representation.

In the third row, all methods struggle to accurately reconstruct the complex roof structures. SD + ControlNet generates more details in the roofs, giving them a more realistic texture. However, it fails to properly reconstruct the roads, with noticeable distortions and misaligned intersections. Satlas-SR and our model handle the roads better, maintaining clearer paths and intersections. Satlas-SR, however, suffers from a lack of contrast and struggles with the reconstruction of small houses, resulting in a more blurred appearance overall. Our model stands out by producing higher contrast and better-defined small structures compared to Satlas-SR, especially in the residential areas with smaller houses. While SD + ControlNet adds realistic texture to some features, the overall balance in road and building reconstruction, combined with superior contrast and clarity, makes our method the closest to the ground truth, especially when considering the finer details in the smaller residential buildings and road layouts.

\begin{table}[!t]
\caption{Quantitative comparison of super-resolution methods on the test set}
\centering
\scalebox{0.95} {
\begin{tabular}{l r r r r}
\toprule
\textbf{Model} & \textbf{PSNR}  & \textbf{SSIM}  & \textbf{LPIPS} $\downarrow$ & \textbf{CLIP}  \\
\midrule
Satlas-SR & \textbf{15.6481} & \textbf{0.2621} & \textbf{0.4786} & 0.9227 \\
SD + ControlNet & 13.4392 & 0.1594 & 0.5640 & 0.8980 \\
Ours & 14.3973 & 0.2221 & 0.4855 & \textbf{0.9261} \\
\bottomrule
\end{tabular}
}
\label{tab:super-resolution-metrics}
\end{table}

\begin{table}[!ht]
    \caption{Segmentation performance of our method and baselines. Our method outperforms others.}
    \centering
    \scalebox{0.9} {
    \begin{tabular}{lccccc}
        \toprule
        \textbf{Model} & \textbf{mIoU}  & \textbf{mF1-score} & \multicolumn{2}{c}{\textbf{Building}} \\
        \cmidrule(lr){4-5}
        &  &  & \textbf{IoU} & \textbf{F1-Score} \\
        \midrule
        Satlas-SR & 51.34 & 59.96 & 20.04 & 29.79\\
        SD + ControlNet & 46.55 & 55.25 & 14.25 & 22.60\\
        Ours & \textbf{55.43} & \textbf{64.04} & \textbf{26.40}  & \textbf{36.79} \\
        \midrule
        NAIP & 59.35 & 67.88 & 32.66 & 43.55 \\
        \bottomrule
    \end{tabular} }
    \label{tab:segmentation-metrics}
\end{table}
\subsection{Segmentation}

For the segmentation task, \cref{tab:segmentation-metrics} reports the mIoU and mF1-score metrics for building segmentation on the super-resolved images. Our model outperforms both Satlas-SR and SD + ControlNet across all segmentation metrics. Specifically, we achieve the highest mIoU of 55.43 and mF1-score of 64.04, significantly surpassing the other models. In terms of building segmentation, our model also performs best with IoU and F1-score values of 26.40 and 36.79, respectively, demonstrating that our super-resolved images provide better foundations for subsequent segmentation tasks. \Cref{fig:segmentation-metrics} provides a qualitative comparison for building segmentation. Our model’s segmentation masks are more accurate and closely resemble the ground truth compared to those generated by Satlas-SR and SD + ControlNet. In both the examples provided, Satlas-SR and SD + ControlNet exhibit issues with boundary detection and misclassification of small buildings, whereas our model provides cleaner and more precise segmentations, especially in densely built environments.

\section{Conclusion}

In this work, we introduced a novel approach to satellite image super-resolution by implementing location guidance, attention mechanisms, and seamless image synthesis techniques. Our method demonstrates state-of-the-art performance, particularly in preserving structural details of things like buildings and roads compared to other models. 

However, we observe the discrepancy between the super-resolution and segmentation metrics. Metrics like PSNR, SSIM, and LPIPS are designed to measure pixel-level fidelity and perceptual quality and do not always correlate with the utility of the super-resolved images in downstream tasks such as segmentation. This suggests that traditional super-resolution metrics may not fully capture the aspects of image quality that are most beneficial for subsequent analytical tasks. These findings underscore the need for a more holistic approach to evaluating super-resolution models in remote-sensing images.

Other than that, challenges related to generalization persist. Although our model excels in standard benchmarks, its ability to generalize to regions significantly different from the training data remains limited. Experiments conducted on areas outside the training set reveal that while our model accurately reconstructs building structures, it struggles to generate new textures or styles, often defaulting to hue changes without adapting underlying textures. This limitation is especially evident when coordinates are shifted to test the model’s adaptability in unfamiliar areas. Full details of this experiment is provided in the supplementary Sec. \textcolor{red}{S5}. These findings indicate that while our method is highly effective in controlled scenarios, improving its ability to adapt to diverse, unseen regions will be crucial for broader real-world applications. Future work will focus on enhancing the generalization capabilities of our model, potentially by incorporating more diverse training datasets, leveraging large generative vision models, or exploring domain adaptation methods to better handle texture variability.

{\small
\bibliographystyle{ieee_fullname}
\bibliography{egbib}
}
\end{document}


\title{Can Location Embeddings Enhance Super-Resolution of Satellite Imagery?
\\ Supplementary Material}

\author{
    \parbox{\textwidth}{\centering
        Daniel Panangian\thanks{Corresponding author}\hspace{1cm}
        Ksenia Bittner \\
        The Remote Sensing Technology Institute \\  
        German Aerospace Center (DLR), Wessling, Germany \\
        \tt\small{\{daniel.panangian, ksenia.bittner\}@dlr.de}
    }
}

\maketitle
\renewcommand{\thesection}{S\arabic{section}}  %
\setcounter{page}{11}
\section{Dataset}
\label{sec:supp-dataset}
In our experiments, we utilize only the RGB channels from Sentinel-2 imagery, simulating true color imagery (TCI) using Sentinel Hub’s L1C Optimized Script\footnote{\href{https://custom-scripts.sentinel-hub.com/sentinel-2/l1c_optimized/}{L1C Optimized Script}}. To refine the training data, we apply a filtering step using WorldCover segmentation labels from the S2-NAIP dataset\cite{satlassuperres}, ensuring that the model is trained on diverse urban environments. Specifically, only tiles with at least 30\% urban coverage are included, resulting in approximately 4,442 tiles in the training set and 95 tiles in the validation set. The test set comprises around 1,883 tiles, selected to evaluate the model’s ability to generalize to regions beyond the United States by incorporating diverse global locations such as forests, mountains, and urban landscapes. Additionally, during inference, we apply the model to various cities outside the United States, as detailed in \cref{sec:supp-results}. This comprehensive evaluation ensures the model's robustness across varying geographical and environmental contexts, which is critical for the building segmentation task.
\section{Implementation Details}
\label{sec:supp-implementation}

\subsection{Baselines and Comparisons}

We use SatLAS-SR\cite{satlassuperres} as our primary baseline, a model that extends ESRGAN\cite{wang2018esrgan} by integrating CLIP loss, an object-aware discriminator, and a feature extractor from a foundation model for remote-sensing super-resolution. SatLAS-SR achieved better results when using multiple Sentinel-2 inputs of the same area taken in different time, but for our experiments, we only use a single Sentinel-2 image that we focus on utilizing location-based features rather than than incorporating temporal data. we also incorporate CLIP loss and the object-aware discriminator, as these components demonstrated the most significant improvements in the paper. To further compare with current state-of-the-art methods, we fine-tune Stable Diffusion\cite{rombach2021high} for the super-resolution task using ControlNet. Stable Diffusion traditionally generates images using text prompts, while ControlNet\cite{zhang2023adding} allows for an additional input, such as an image, to guide the generation process. This conditioning mechanism enables greater control over the output by leveraging input images like edges, depth maps, or other structural cues, which the model uses alongside the text prompt to shape the generated image. For a fair comparison, we fine-tune Stable Diffusion on our dataset and use a dummy text prompt for each sample to simulate similar input conditions across all models. Given that Stable Diffusion operates on 512×512 pixel images, we first upsample both Sentinel and NAIP imagery to 512×512 before downscaling them back to their original resolution after processing.

\subsection{Training Details}
For all experiments, our model is implemented in PyTorch and runs on an NVIDIA A100-SXM4 GPU with a batch size of 16. For training hyperparameters, we follow Satlas-SR\cite{satlassuperres}. All models are trained from scratch using the Adam optimizer, with the learning rate initialized to $10^{-4}$. For both the generator and discriminator, we employ the \textit{large} variant mentioned in the SatLAS-SR paper, which includes 256 features, 128 grow channels, and 30 blocks. For our location-matching discriminator and location embedding features, we use 64 features. The loss function is a weighted combination of several components: the pixel loss \( \lambda_{\text{pix}} = 1.0 \), perceptual loss \( \lambda_{\text{perceptual}} = 1.0 \), GAN loss \( \lambda_{\text{GAN}} = 0.1 \), CLIP loss \( \lambda_{\text{CLIP}} = 1.0 \), OpenStreetMap loss \( \lambda_{\text{OSM}} = 0.3 \), and our location-matching loss \( \lambda_{\text{loc\_match}} = 1.0 \).

\section{Qualitative results}
\label{sec:supp-results}
\begin{figure*}[!ht]
\begin{center}
\includegraphics[width=17cm]{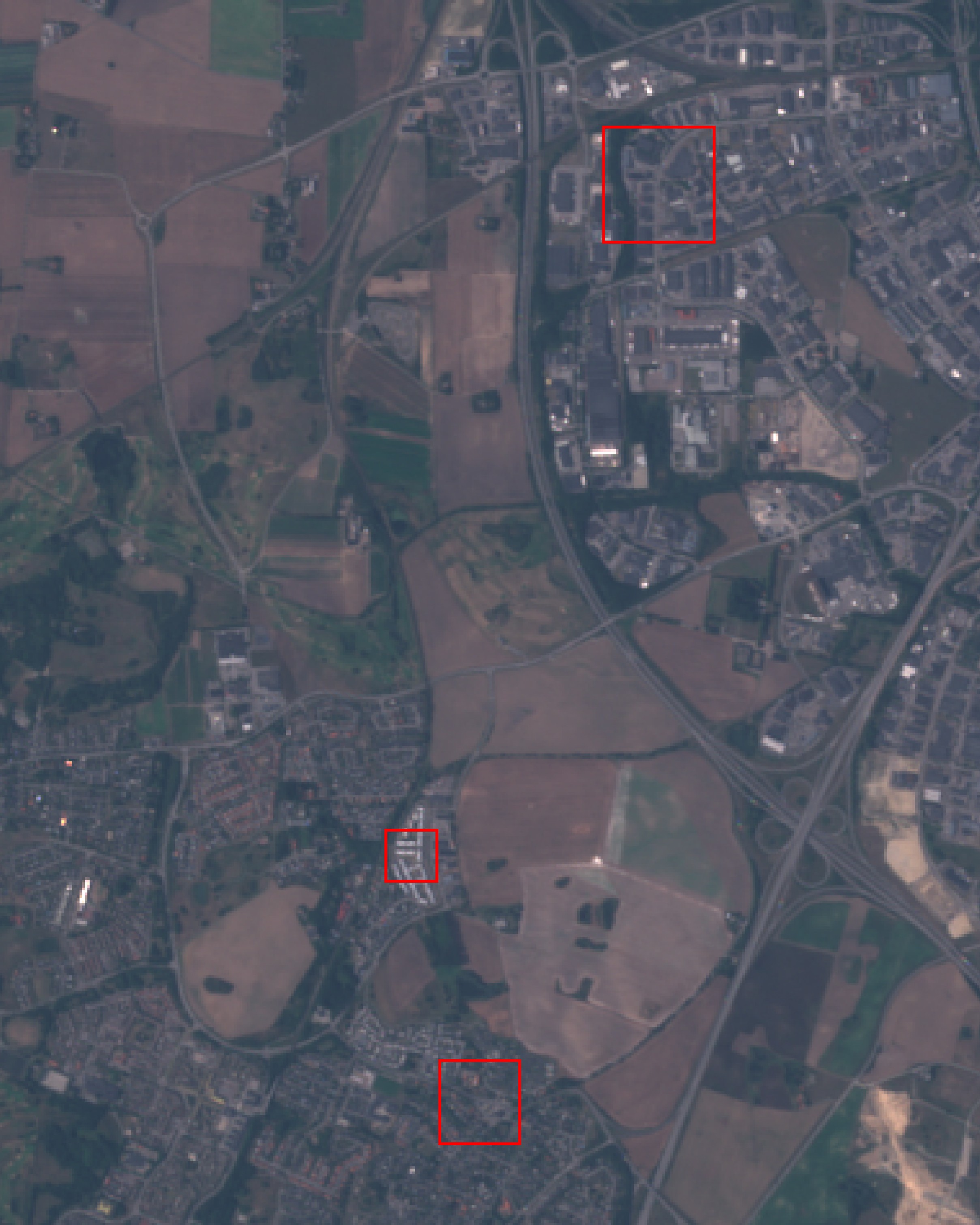}
\end{center}
\caption{Sentinel-2 input image of the Malmö region}

\label{fig:copenhagen-sentinel}
\end{figure*}

\begin{figure*}[!ht]
\begin{center}
\includegraphics[width=17cm]{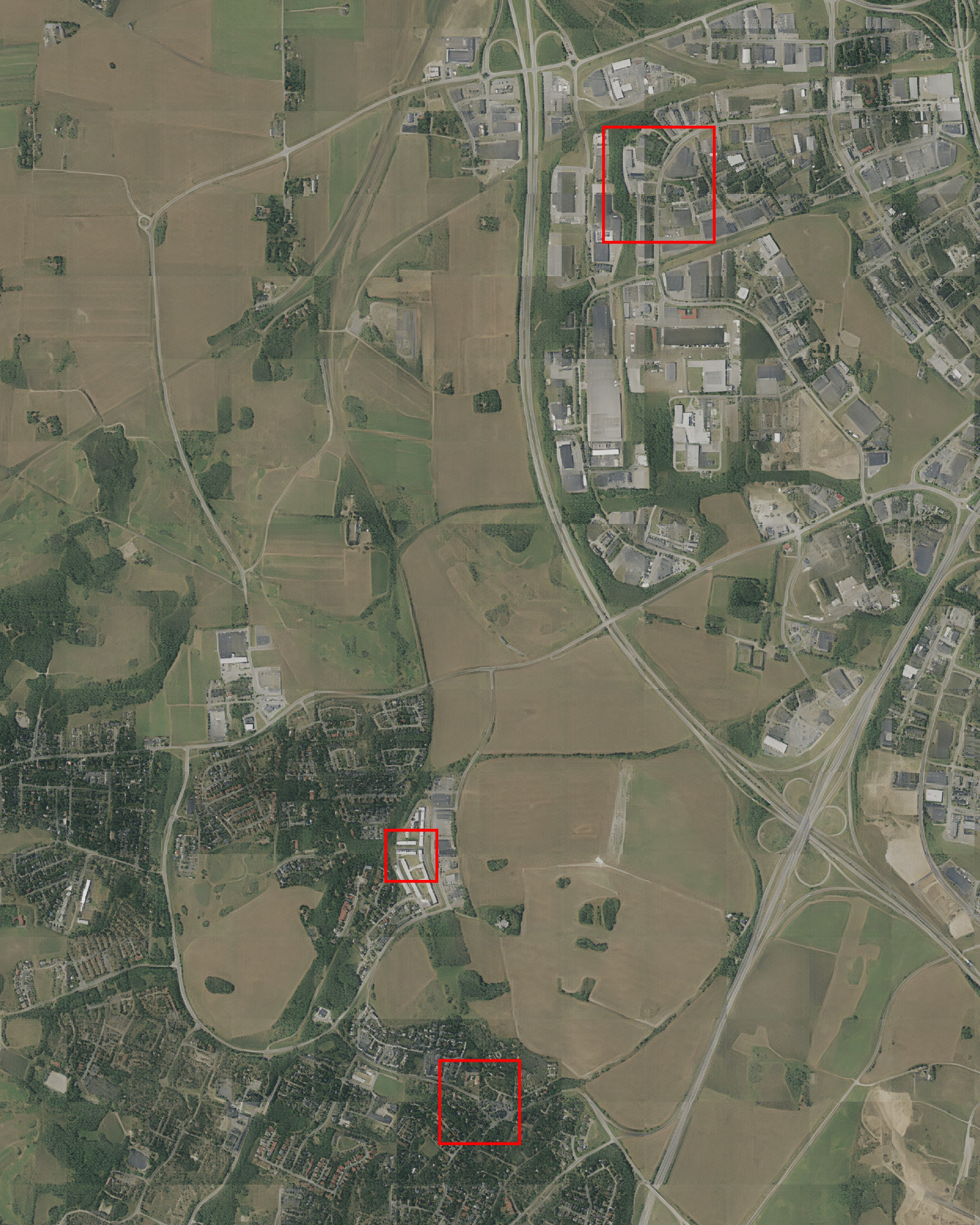}
\end{center}
\caption{Super-resolution output from Satlas-SR for the Malmö region}

\label{fig:copenhagen-satlas}
\end{figure*}

\begin{figure*}[!ht]
\begin{center}
\includegraphics[width=17cm]{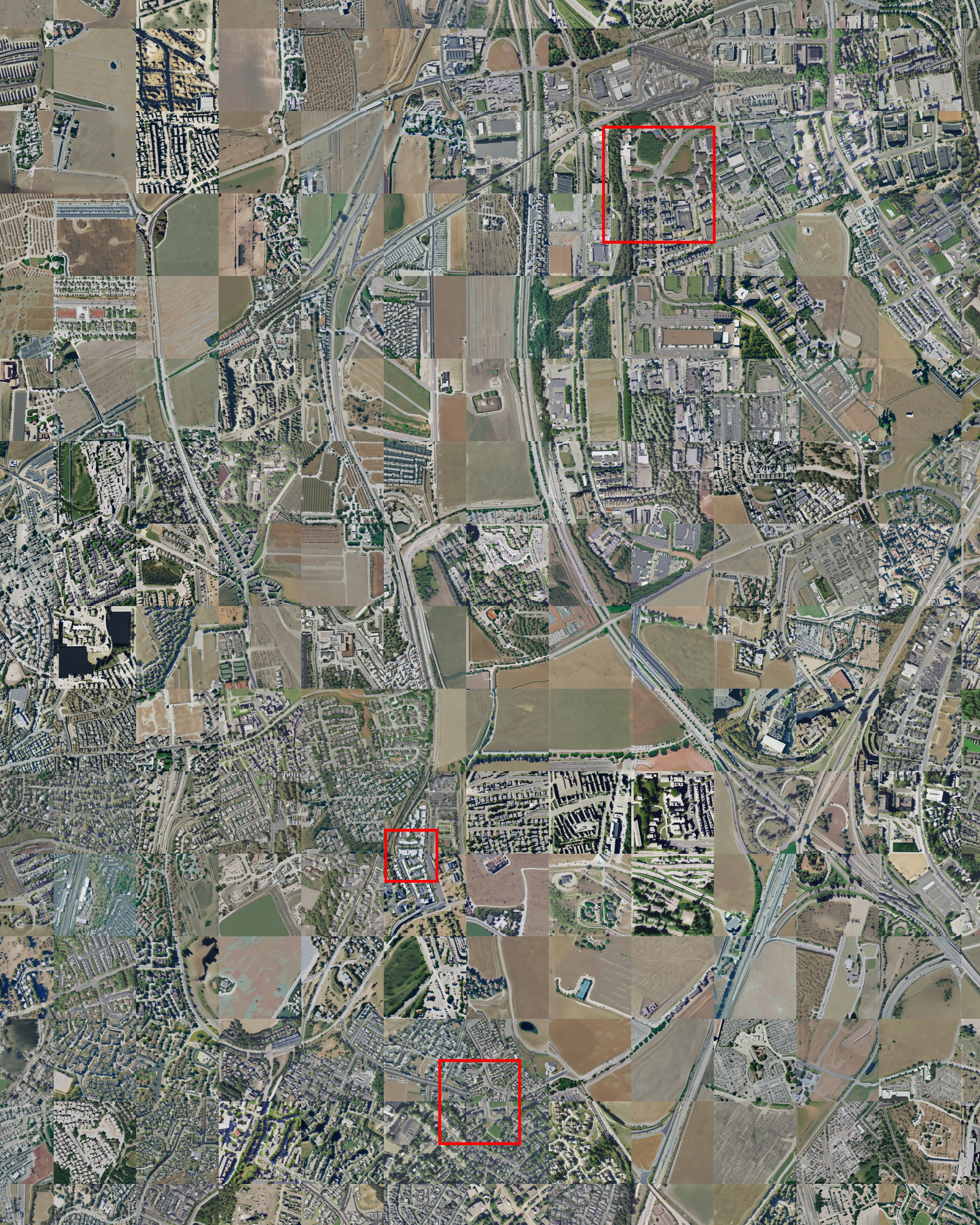}
\end{center}
\caption{Super-resolution output from Stable Diffusion with ControlNet for the Malmö region}
\label{fig:copenhagen-sd}
\end{figure*}

\begin{figure*}[!ht]
\begin{center}
\includegraphics[width=17cm]{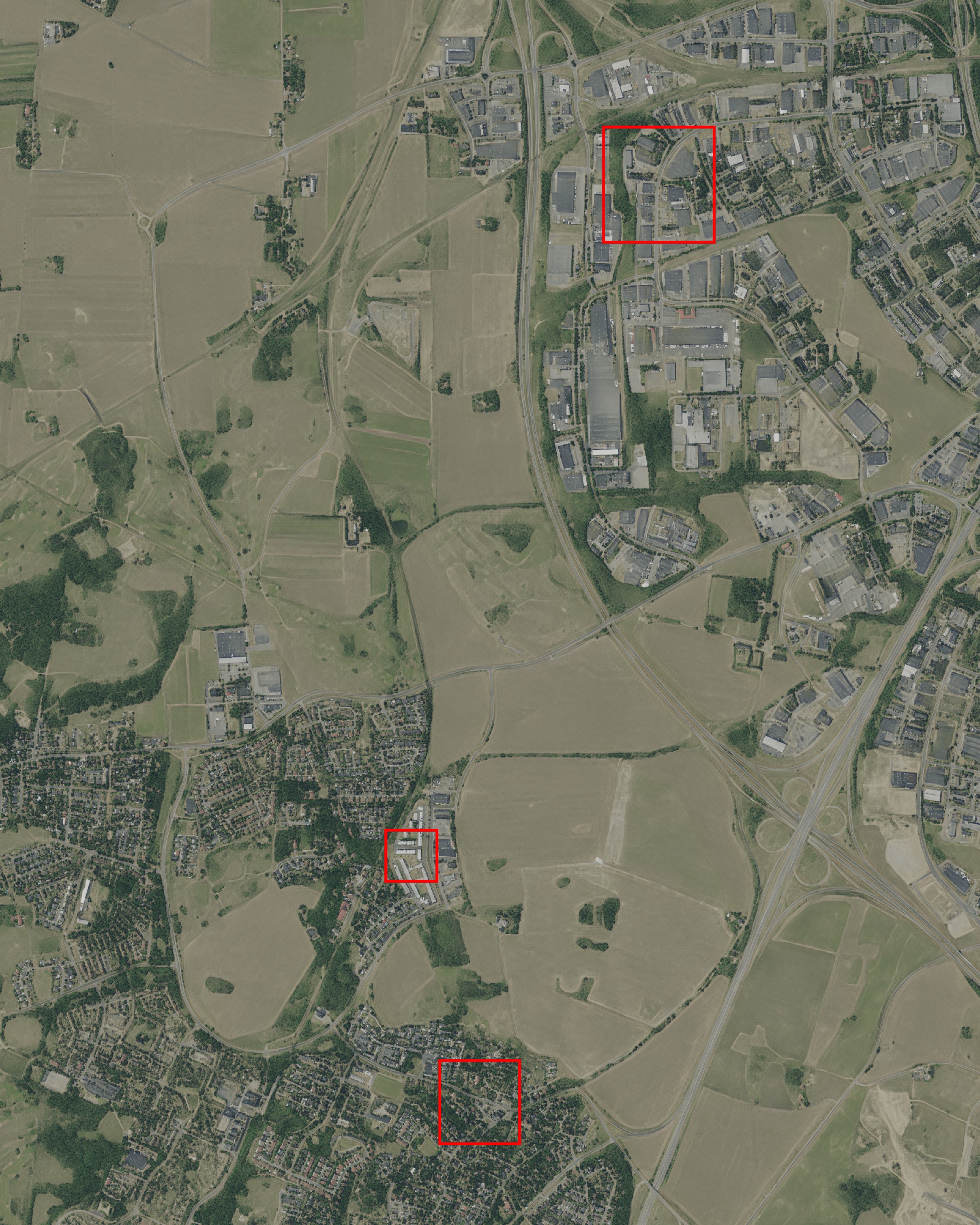}
\end{center}
\caption{Super-resolution output from our method for the Malmö region
}
\label{fig:copenhagen-ours}
\end{figure*}

We performed super-resolution inference over a large area in Malmö, Denmark to assess the performance of different methods at scale. The comparison includes four sets of figures: the low-resolution Sentinel-2 input (Fig. \ref{fig:copenhagen-sentinel}), outputs from Satlas-SR (Fig. \ref{fig:copenhagen-satlas}), Stable Diffusion (SD) + ControlNet (Fig. \ref{fig:copenhagen-sd}), and our method (Fig. \ref{fig:copenhagen-ours}). Starting with SD + ControlNet, while it offers some improvement in generating finer details, it suffers from significant inconsistencies in both color and texture across patches. These inconsistencies become especially problematic when stitching the patches together, resulting in a disjointed appearance with noticeable blocky patterns. The overall image looks fragmented, as if each patch comes from a completely different area. This lack of cohesion is particularly disruptive in regions with homogeneous textures, such as agricultural fields or open urban areas, where smooth transitions are expected. In the other hand, Satlas-SR and our method show much more consistent color and texture across patches. However, Satlas still displays subtle block patterns. These patterns arise because the method does not share sufficient context between patches, leading to noticeable tiling effects in areas where the texture is expected to remain constant. Our method, on the other hand, effectively overcomes this limitation by ensuring smooth transitions between patches, preserving both color and texture consistency throughout the entire scene. This leads to a more visually cohesive result, without the blocky artifacts seen in the other methods.

\begin{figure*}[htbp]
    \centering

    \begin{subfigure}[b]{0.25\textwidth}
        \centering
        \includegraphics[height=4cm]{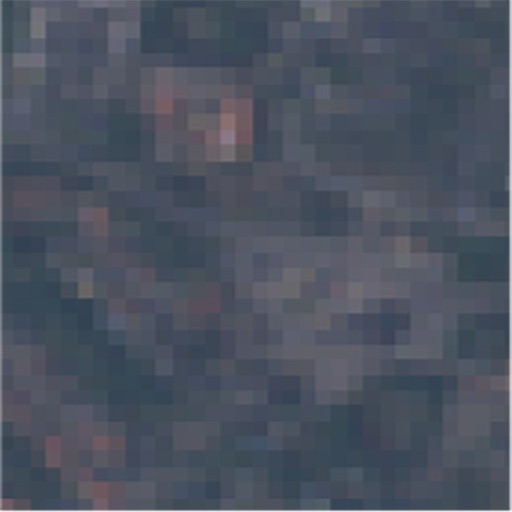} %
    \end{subfigure}
    \begin{subfigure}[b]{0.25\textwidth}
        \centering
        \includegraphics[height=4cm]{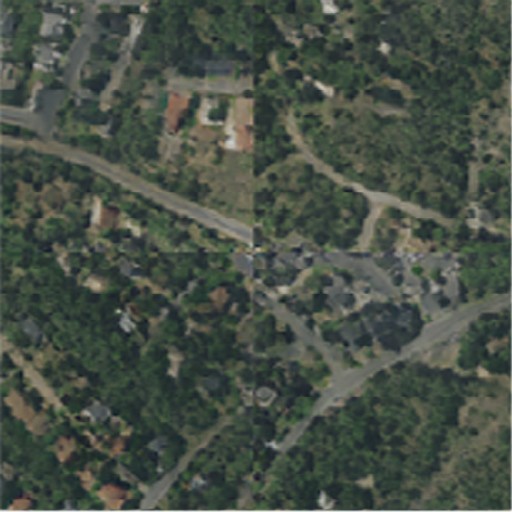} %
    \end{subfigure}
    \begin{subfigure}[b]{0.25\textwidth}
        \centering
        \includegraphics[height=4cm]{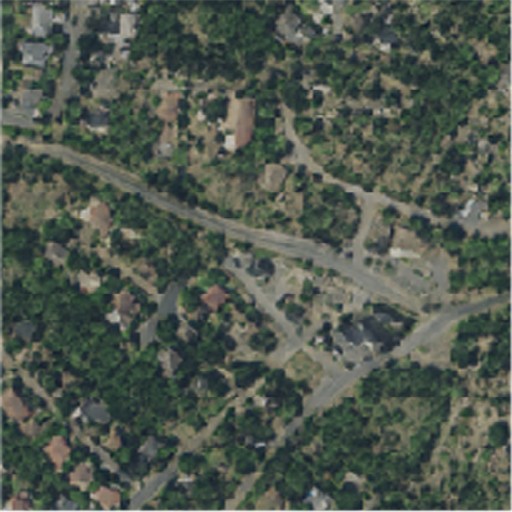} %
    \end{subfigure}

    \vspace{1em} %

    \begin{subfigure}[b]{0.25\textwidth}
        \centering
        \includegraphics[height=4cm]{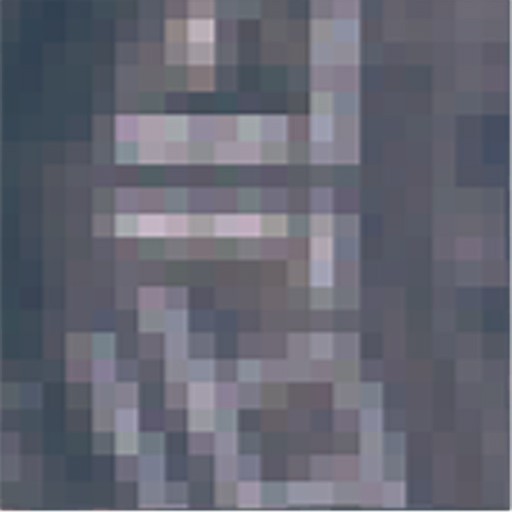} %
    \end{subfigure}
    \begin{subfigure}[b]{0.25\textwidth}
        \centering
        \includegraphics[height=4cm]{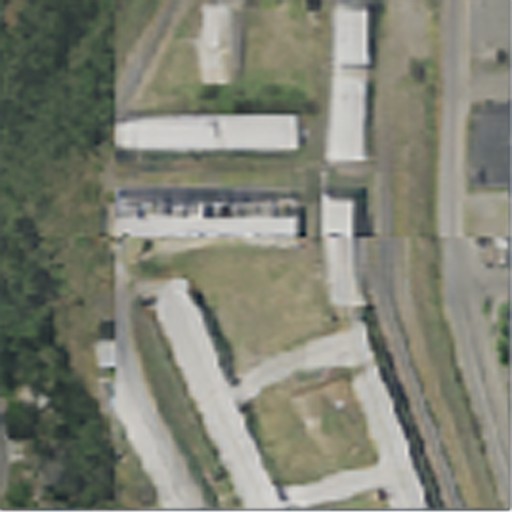} %
    \end{subfigure}
    \begin{subfigure}[b]{0.25\textwidth}
        \centering
        \includegraphics[height=4cm]{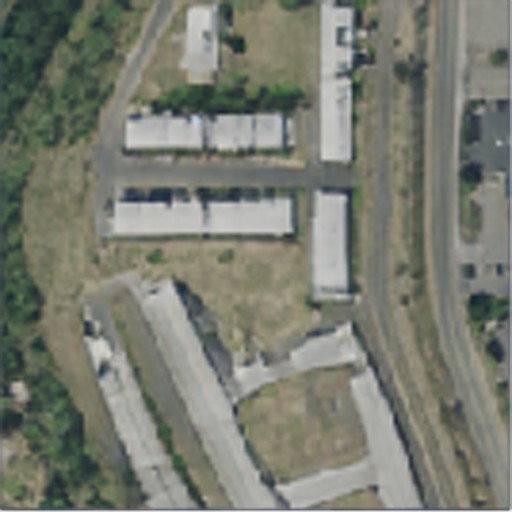} %
    \end{subfigure}

    \vspace{1em} %
    \begin{subfigure}[b]{0.25\textwidth}
        \centering
        \includegraphics[height=4cm]{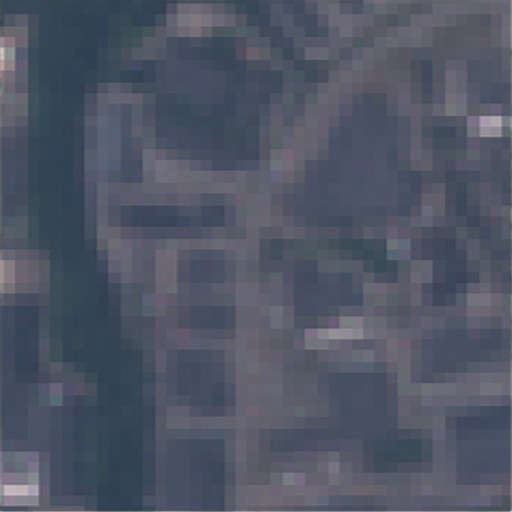} %
        \caption{Sentinel-2}
    \end{subfigure}
    \begin{subfigure}[b]{0.25\textwidth}
        \centering
        \includegraphics[height=4cm]{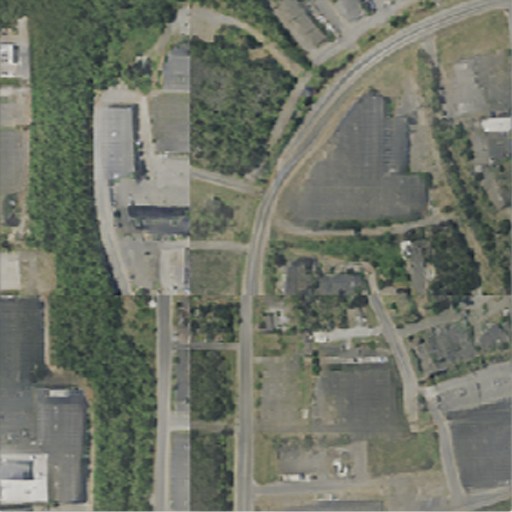} %
        \caption{Satlas-SR}
    \end{subfigure}
    \begin{subfigure}[b]{0.25\textwidth}
        \centering
        \includegraphics[height=4cm]{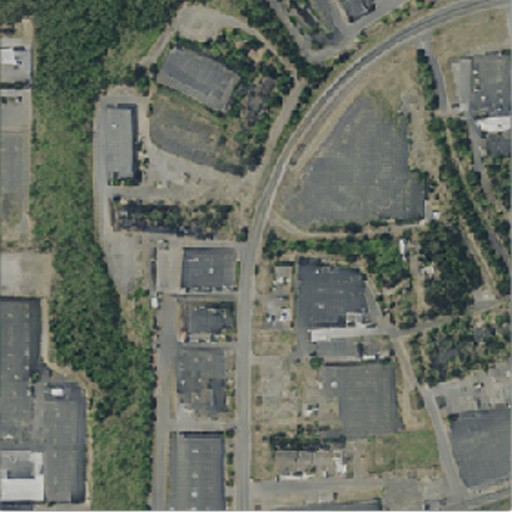} %
        \caption{Ours}
    \end{subfigure}
    \caption{Comparison of super-resolution outputs for regions of interest from the Malmö inference}
    \label{fig:local_padding}
\end{figure*}

A closer inspection reveals several challenges in applying super-resolution independently to each patch, as seen in the results from Satlas-SR and ours in \ref{fig:local_padding}. For example, in urban residential areas, we observe that roads are often interrupted between patches, creating discontinuities that disrupt the visual flow of the scene. Similarly, in industrial zones, large structures like buildings can become incomplete, with parts missing in certain patches. In suburban neighborhoods, several buildings appear fragmented or distorted, breaking the continuity of the image. Our method addresses these inconsistencies by using information from neighboring patches to maintain coherence across the entire area.

\section{Ablation study}
\label{sec:supp-ablation}
 We analyze the impact of adding self-attention and location embeddings (via cross-attention) on the model's performance. We compare four configurations: a baseline model (Satlas-SR), a model with self-attention, a model with both self-attention and location embeddings, and the final model, which includes a location matching discriminator. The performance metrics for these models, including PSNR, SSIM, LPIPS, and CLIP Score, are shown in Table \ref{tab:ablation}.

The baseline model exhibits balanced performance across metrics. Adding self-attention in the model leads to improvements, with slightly higher PSNR, SSIM, and CLIP scores, indicating better image quality and semantic alignment. However, the inclusion of location embeddings results in a trade-off. While the PSNR and SSIM drop to 14.5957 and 0.2453 respectively, the CLIP score improves to 0.9373. Finally, the addition of the location matching discriminator in the final model slightly decreases all metrics although the CLIP score remains relatively high at 0.9261.

This results reveals that while the addition of location embeddings and a location matching discriminator enhances semantic alignment, it introduces trade-offs in traditional image quality metrics.

\begin{table}[!h]
\centering
\small %
\scalebox{0.9}{
\begin{tabular}{p{0.30\columnwidth} r r r r} %
\toprule
\textbf{Model} & \textbf{PSNR} & \textbf{SSIM} & \textbf{LPIPS} & \textbf{CLIP Score} \\
\midrule
ESRGAN &  & &  \\
\hspace{10pt}+ CLIP+ OSM & 15.6481 & 0.2621 & 0.4786 & 0.9227 \\
\midrule
\textbf{\hspace{10pt} + Self-Att.} & 15.7768 & 0.2663 & 0.4624 & 0.9334 \\
\textbf{\hspace{10pt} + Location} & 14.5957 & 0.2453 & 0.4668 & 0.9373 \\
\textbf{\hspace{10pt} + Loc. Disc} & 14.3973 & 0.2221 & 0.4855 & 0.9261 \\
\bottomrule
\end{tabular}}
\caption{Ablation study results of different model configurations.}
\label{tab:ablation}
\end{table}
\section{Location control}
\label{sec:supp-control}
\begin{figure*}[!t]
    \centering
    \begin{subfigure}[t]{0.23\textwidth}
        \centering
        \textbf{Original}\\ \hfill
        \\
        \vspace{0.5em} %
        \includegraphics[width=\textwidth]{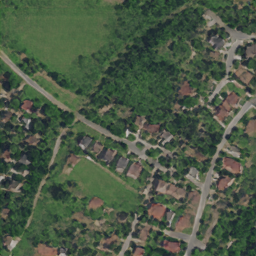} \\  %
        \vspace{0.5em}
        \includegraphics[width=\textwidth]{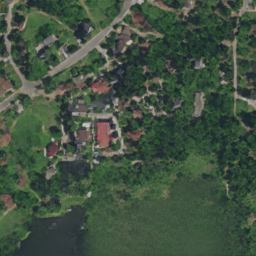} \\  %
        \vspace{0.5em}
        \includegraphics[width=\textwidth]{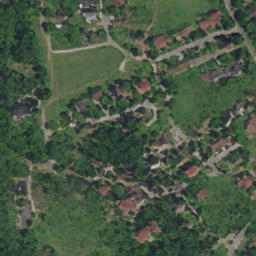}      %
    \end{subfigure}
    \hfill
    \begin{subfigure}[t]{0.23\textwidth}
        \centering
        \textbf{Seattle,}\\ \textbf{Washington}\\
        \vspace{0.5em}
        \includegraphics[width=\textwidth]{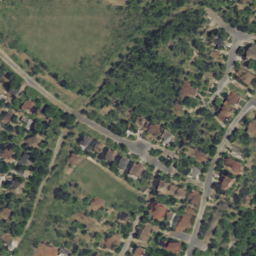} \\  %
        \vspace{0.5em}
        \includegraphics[width=\textwidth]{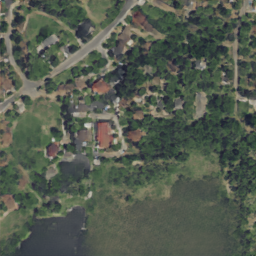} \\  %
        \vspace{0.5em}
        \includegraphics[width=\textwidth]{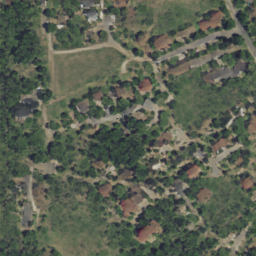}      %
    \end{subfigure}
    \hfill
    \begin{subfigure}[t]{0.23\textwidth}
        \centering
        \textbf{San Diego,}\\ \textbf{California}\\
        \vspace{0.5em}
        \includegraphics[width=\textwidth]{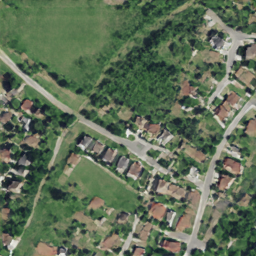} \\  %
        \vspace{0.5em}
        \includegraphics[width=\textwidth]{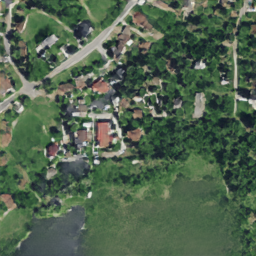} \\  %
        \vspace{0.5em}
        \includegraphics[width=\textwidth]{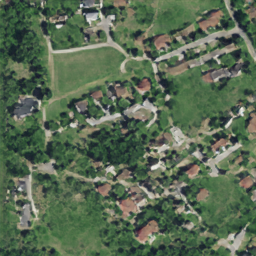}      %
    \end{subfigure}
    \hfill
    \begin{subfigure}[t]{0.23\textwidth}
        \centering
        \textbf{Des Moines,}\\ \textbf{Iowa}\\
        \vspace{0.5em}
        \includegraphics[width=\textwidth]{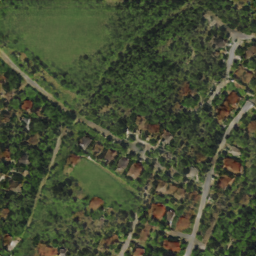} \\  %
        \vspace{0.5em}
        \includegraphics[width=\textwidth]{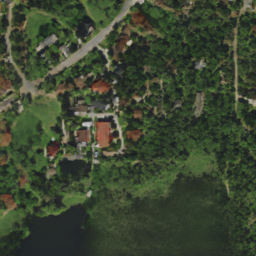} \\  %
        \vspace{0.5em}
        \includegraphics[width=\textwidth]{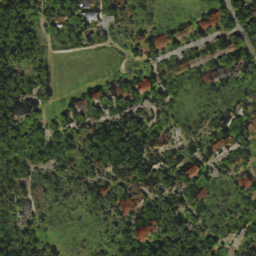}      %
    \end{subfigure}
    \hfill

    \caption{Super-resolution results using different geographic coordinates. The first column shows the output when using the original coordinates from an area in Germany. The subsequent columns show the results of applying super-resolution with different coordinates in USA.}
    \label{fig:location_control}

\end{figure*}

In this section, we evaluate the performance of the location control feature in our model, which incorporates geographic coordinates into the super-resolution process. To test this feature, we applied SR to the same input image while varying the coordinates used in the location embeddings. We compared the results generated using the original coordinates of the input image with outputs produced using coordinates from different cities in USA, as shown in Figure \ref{fig:location_control}. These locations were specifically chosen for their distinct geographic and environmental characteristics. While the model effectively maintains the overall content of the images (i.e., shapes, textures, and spatial arrangement of objects), the only noticeable change when altering the coordinates is in the hue of the images. The structural elements, such as building shapes and road layouts, remain consistent across different locations. However, this limited change in image characteristics reveals a limitation of the current location control mechanism: it does not fully capture the regional variations that would be expected when changing geographic context. These variations could include differences in architectural styles, natural features, and other location-specific characteristics, which are not reflected in the generated images.

{\small
\bibliographystyle{ieee_fullname}
\bibliography{egbib}
}